\pdfoutput=1

\documentclass[11pt]{article}

\usepackage{ACL2023}

\usepackage{times}
\usepackage{latexsym}
\usepackage{url}
\usepackage{amsmath}
\usepackage{adjustbox}
\usepackage[T1]{fontenc}
\usepackage[yyyymmdd]{datetime}

\usepackage[utf8]{inputenc}
\usepackage{textcomp}
\usepackage{microtype}

\usepackage{inconsolata}
\usepackage{tcolorbox}
\usepackage{enumitem}
\usepackage{cleveref}
\usepackage{booktabs}
\usepackage{longtable}
\crefname{section}{§}{§§}
\Crefname{section}{§}{§§}

\usepackage{xspace}

\definecolor{lightgrey}{rgb}{0.95, 0.95, 0.95}

\newcommand{\ours}[0]{NeuSym-RAG\xspace}
\newcommand{\dataset}[0]{\textsc{AirQA-Real}\xspace}
\newcommand{\retrievefromdatabase}[0]{\textsc{RetrieveFromDatabase}\xspace}
\newcommand{\retrievefromvectorstore}[0]{\textsc{RetrieveFromVectorstore}\xspace}
\newcommand{\calculateexpr}[0]{\textsc{CalculateExpr}\xspace}
\newcommand{\viewimage}[0]{\textsc{ViewImage}\xspace}
\newcommand{\generateanswer}[0]{\textsc{GenerateAnswer}\xspace}
\newcommand{\numpdf}[0]{$6797$\xspace}
\newcommand{\numquestion}[0]{$553$\xspace}
\newcommand{\numannotator}[0]{$16$\xspace}
\newcommand{\gap}[0]{$17.3\%$\xspace}
\newcommand{\metricsize}[0]{$18$\xspace}
\usepackage{multicol}
\usepackage{multirow}
\usepackage{listings}
\usepackage{caption}
\usepackage{fancyvrb}
\definecolor{delim}{RGB}{20,105,176}
\definecolor{numb}{RGB}{106, 109, 32}
\definecolor{string}{rgb}{0.64,0.08,0.08}
\definecolor{backcolour}{rgb}{0.95,0.95,0.92}

\definecolor{pykeyword}{RGB}{0, 0, 139}     
\definecolor{pystring}{RGB}{165, 42, 42}  
\definecolor{pycomment}{RGB}{85, 107, 47}  
\definecolor{pynumber}{RGB}{105, 105, 105}
\definecolor{darkred}{rgb}{0.5,0,0}
\lstdefinelanguage{json}{
    numbers=left,
    numberstyle=\small,
    frame=single,
    rulecolor=\color{black},
    showspaces=false,
    showstringspaces=false,
    showtabs=false,
    breaklines=true,
    postbreak=\raisebox{0ex}[0ex][0ex]{\ensuremath{\color{gray}\hookrightarrow\space}},
    breakatwhitespace=true,
    basicstyle=\ttfamily\small,
    upquote=true,
    morestring=[b]",
    stringstyle=\color{string},
    literate=
     *{0}{{{\color{numb}0}}}{1}
      {1}{{{\color{numb}1}}}{1}
      {2}{{{\color{numb}2}}}{1}
      {3}{{{\color{numb}3}}}{1}
      {4}{{{\color{numb}4}}}{1}
      {5}{{{\color{numb}5}}}{1}
      {6}{{{\color{numb}6}}}{1}
      {7}{{{\color{numb}7}}}{1}
      {8}{{{\color{numb}8}}}{1}
      {9}{{{\color{numb}9}}}{1}
      {\{}{{{\color{delim}{\{}}}}{1}
      {\}}{{{\color{delim}{\}}}}}{1}
      {[}{{{\color{delim}{[}}}}{1}
      {]}{{{\color{delim}{]}}}}{1},
}
\usepackage{amssymb}
\usepackage{bbding}
\usepackage{enumitem}
\usepackage{makecell}
\usepackage{tikz}
\usepackage{arydshln}
\usepackage{textcomp}

\definecolor{iterneurag}{RGB}{251, 231, 163}
\definecolor{itersymrag}{RGB}{194, 214, 236}
\definecolor{neusymrag}{RGB}{170, 220, 170}
\definecolor{wordcolor}{RGB}{242, 242, 242}

\title{\ours: Hybrid Neural Symbolic Retrieval with Multiview Structuring for PDF Question Answering}

\author{Ruisheng Cao$^{12}$\thanks{\ \ Equal contribution.}, Hanchong Zhang$^{12*}$, Tiancheng Huang$^{12*}$, Zhangyi Kang$^{12}$, Yuxin Zhang$^{12}$,\\
\textbf{Liangtai Sun$^{12}$, Hanqi Li$^{124}$, Yuxun Miao$^{12}$, Shuai Fan$^{23}$, Lu Chen$^{124\dagger}$ and Kai Yu$^{124}$\thanks{\ \ The corresponding authors are Lu Chen and Kai Yu.}}\\
  $^{1}$MoE Key Lab of Artificial Intelligence, Shanghai, China\\
  X-LANCE Lab, School of Computer Science, Shanghai Jiao Tong University, Shanghai, China\\
  $^{2}$Jiangsu Key Lab of Language Computing, Suzhou, China\\
  $^{3}$AISpeech Co., Ltd., Suzhou, China\quad $^{4}$Suzhou Laboratory, Suzhou, China\\
  {\tt \{211314,htc981,narcisss,chenlusz,kai.yu\}@sjtu.edu.cn}\\}

\begin{document}
\maketitle

\begin{abstract}
The increasing number of academic papers poses significant challenges for researchers to efficiently acquire key details.
While retrieval augmented generation (RAG) shows great promise in large language model (LLM) based automated question answering,
previous works often isolate neural and symbolic retrieval despite their complementary strengths.
Moreover, conventional single-view chunking neglects the rich structure and layout of PDFs, e.g., sections and tables.
In this work, we propose \ours, a hybrid neural symbolic retrieval framework which combines both paradigms in an interactive process. By leveraging multi-view chunking and schema-based parsing, \ours organizes semi-structured PDF content into both the relational database and vectorstore, enabling LLM agents to iteratively gather context until sufficient to generate answers.
Experiments on three full PDF-based QA datasets, including a self-annotated one \dataset, show that \ours stably defeats both the vector-based RAG and various structured baselines, highlighting its capacity to unify both retrieval schemes and utilize multiple views.
\end{abstract}
\section{Introduction}
With the exponential growth in academic papers, large language model~(LLM) based question answering~(QA) systems show great potential to help researchers extract key details from emerging studies.
However, individual PDFs often exceed prompt limits, and user queries may span multiple documents.
To tackle these challenges, retrieval-augmented generation~(RAG,~\citealp{rag-survey}) is effective for knowledge-intensive QA.

\begin{figure}[htbp]
    \centering
    \includegraphics[width=0.5\textwidth]{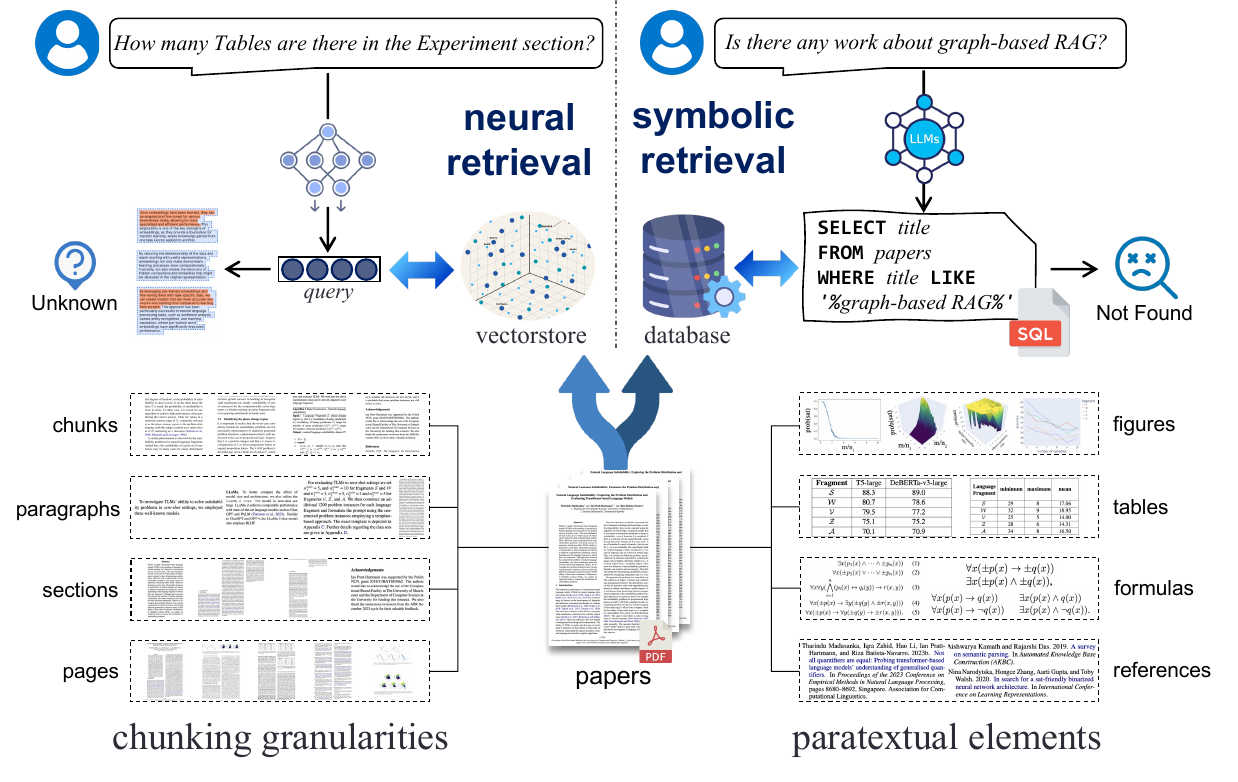}
    \caption{Motivation of the proposed \ours. (Upper) Two paradigms of retrieval strategies. (Bottom) PDF documents can be split based on different granularities and they contain many paratextual elements.}
    \label{fig:motivation}
\end{figure}

Despite its wide application, the classic \emph{neural retrieval}~\cite{realm} often fails when handling precise queries involving mathematical operations, comparisons, or aggregations. For example, in the top-left of Figure~\ref{fig:motivation}, the total number of tables cannot be determined through retrieved chunks as they are scattered across the document. On the other hand, \emph{symbolic retrieval} such as TAG~\cite{tag} relies on semantic parsing~\cite{semantic-parsing} techniques, e.g., text-to-SQL~\cite{birdsql}, to directly extract the target information from the structured database.
Unfortunately, such precise queries often break down in semantic fuzzy matching or morphological variations, e.g., ``\emph{graph-based RAG}'' versus ``\emph{GraphRAG}''.
Previous literature investigates these two paradigms in isolation.

Furthermore, the most widely utilized scheme to segment documents into chunks is based on a fixed length of consecutive tokens, possibly considering sentence boundaries or more delicate granularities~\cite{small2big}.
However, for semi-structured PDF documents of research papers, this common practice neglects the intrinsic structure of sections and the salient features of paratextual tables and figures~(illustrated at the bottom of Figure~\ref{fig:motivation}). The distinct layout of PDF files offers a more structured view towards segmenting and arranging content.

To this end, we propose a hybrid \textbf{Neu}ral \textbf{Sym}bolic retrieval framework (\ours) for PDF-based question answering, which combines both retrieval paradigms into an interactive procedure. During pre-processing, the PDF file of each paper is fed into a pipeline of parsing functions~(\cref{sec:pdf_parsing}) to complete the metadata, segment raw texts based on different views, and extract various embedded paratextual elements~(e.g., tables and figures). These identified elements are populated into a relational database, specifically designed for semi-structured PDF. Then, we select encodable column values from the populated database for storage into the vectorstore~(\cref{sec:multimoda-encoding}). These cell values present diverse views in interpreting PDF content. Moreover, the database schema graph is leveraged to organize the vectors into a well-formed structure. To answer the user question, LLM agents can adaptively predict executable actions~(\cref{sec:actions}) to retrieve desired information from either backend environment~(database or vectorstore) in a multi-round fashion. This agentic retrieval terminates when the collected information suffices to answer the input question, in which case LLM agent will predict a terminal ``\generateanswer'' action.

To validate \ours, we convert and annotate three full PDF-based academic QA datasets, including a self-curated one \dataset with \numquestion samples and \metricsize instance-specific evaluation metrics~(\cref{sec:dataset}).
Experiments on both closed- and open-source LLMs demonstrate that it remarkably outperforms the classic neural RAG~(\gap points on \dataset) and various structured methods like HybridRAG~\cite{hybridrag} and GraphRAG~\cite{graphrag}. 
The ablation study highlights: 1) the positive impact of model size increase on agentic retrieval, 2) the superiority of integrating multiple chunking perspectives, and 3) the low sensitivity to LLM choice for pre-processing and high sensitivity for subjective evaluation.

To sum up, our contributions are three-fold:
\begin{itemize}
    \item We are the first to integrate both vector-based neural retrieval and SQL-based symbolic retrieval into a unified and interactive \ours framework through executable actions.
    \item We incorporate multiple views for parsing and vectorizing PDF documents, and adopt a structured database schema to systematically organize both text tokens and encoded vectors.
    
    \item Experiments on three realistic full PDF-based QA datasets \emph{w.r.t.} academic research
    validate the superiority of \ours over various neural and symbolic baselines.
\end{itemize}

\section{Framework of NeuSym-RAG}
This section presents the complete framework of \ours, comprising three stages~(Figure~\ref{fig:framework}).

\subsection{The Overall Procedure}

To answer user questions about a semi-structured PDF file, the entire workflow proceeds as follows:
\begin{figure*}[htbp]
    \centering
    \includegraphics[width=0.95\textwidth]{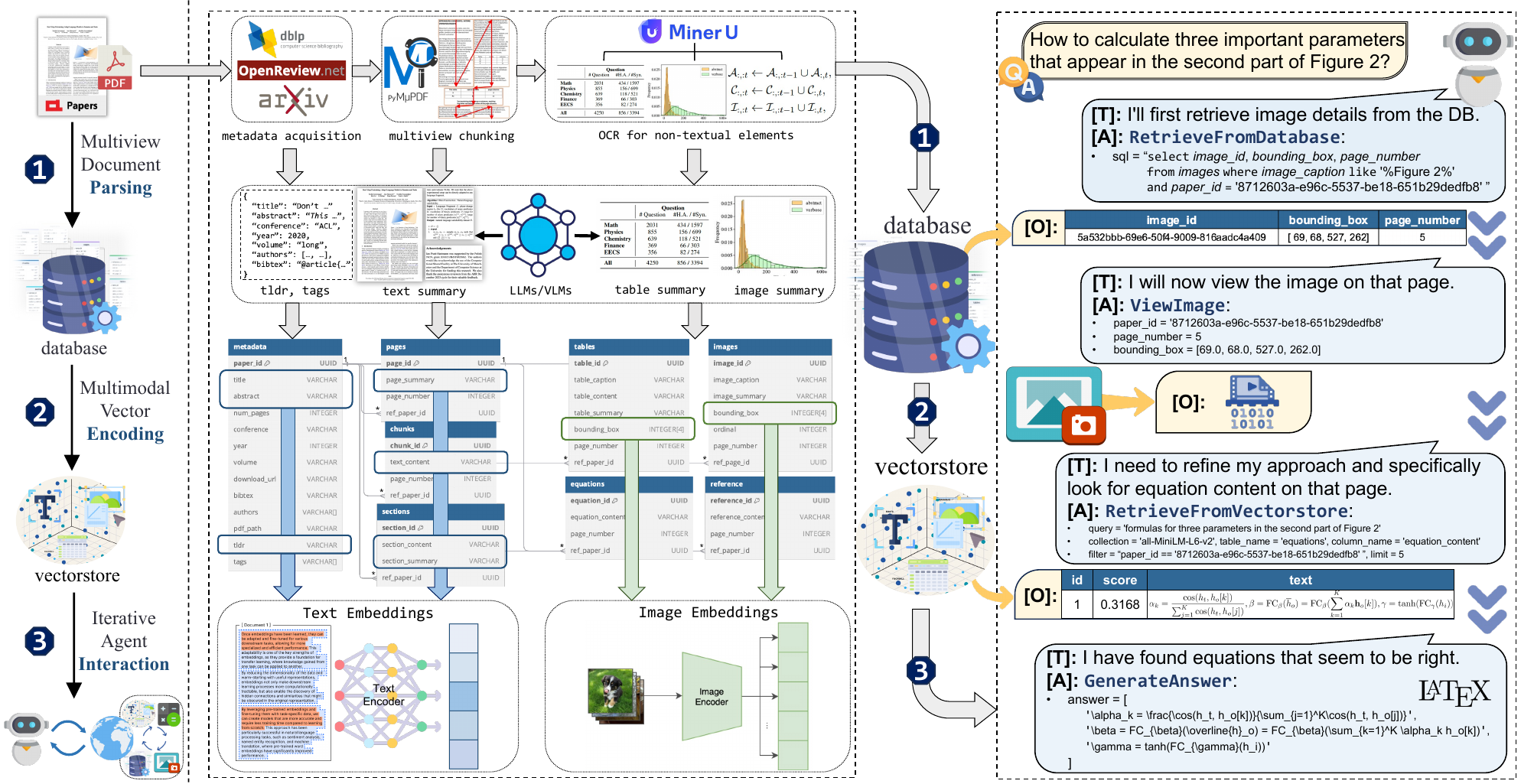}
    \caption{Overview of the \ours framework. The demonstration example comes from a real and simplified case in our labeled dataset~(\cref{sec:dataset}). ``[T]'', ``[A]'', and ``[O]'' represent thought, action and observation, respectively.}
    \label{fig:framework}
\end{figure*}

\begin{enumerate}[label=\arabic*),leftmargin=*]
    \item \textbf{Parsing:} Firstly, we pass the raw PDF file into a pipeline of functions to segment it in multi-view, extract non-textual elements, and store them in a schema-constrained database~(DB).
    \item \textbf{Encoding:} Next, we identify those encodable columns in the DB, and utilize embedding models for different modalities to obtain and insert vectors of cell values into the vectorstore~(VS). 
    \item \textbf{Interaction:} Finally, we build an iterative Q\&A agent which can predict executable actions to retrieve context from the backend environment~(either DB or VS) and answer the input question.
\end{enumerate}

\subsection{Multiview Document Parsing}
\label{sec:pdf_parsing}
At this stage, for each incoming PDF file, we aim to parse it with different perspectives into a relational database DuckDB~\cite{duckdb}. The pipeline of parsing functions includes~(in the top middle of Figure~\ref{fig:framework}): 1) Querying scholar APIs~(e.g., \texttt{arxiv}) 
to obtain the metadata such as the authors and published conference, such that we can support metadata-based filtering~\cite{meta-rag} during retrieval. 2) Splitting the text based on different granularities with tool PyMuPDF~\cite{pymupdf}, e.g., pages, sections, and fixed-length continuous tokens. 3) Leveraging OCR models to extract non-textual elements~(we choose the tool \texttt{MinerU},~\citealp{mineru}). 4) Asking large language models~(LLMs) or vision language models~(VLMs) to generate concise summaries of the parsed texts, tables, and images~\cite{recomp}. The retrieved metadata, parsed elements and predicted summaries will all be populated into the symbolic DB. We handcraft the DB schema in advance, which is carefully designed and universal for PDF documents~(see the middle part of Figure~\ref{fig:framework} or App.~\ref{app:database_schema} for visualization).

\subsection{Multimodal Vector Encoding}
\label{sec:multimoda-encoding}
\begin{figure}[htbp]
    \centering
    \includegraphics[width=0.45\textwidth]{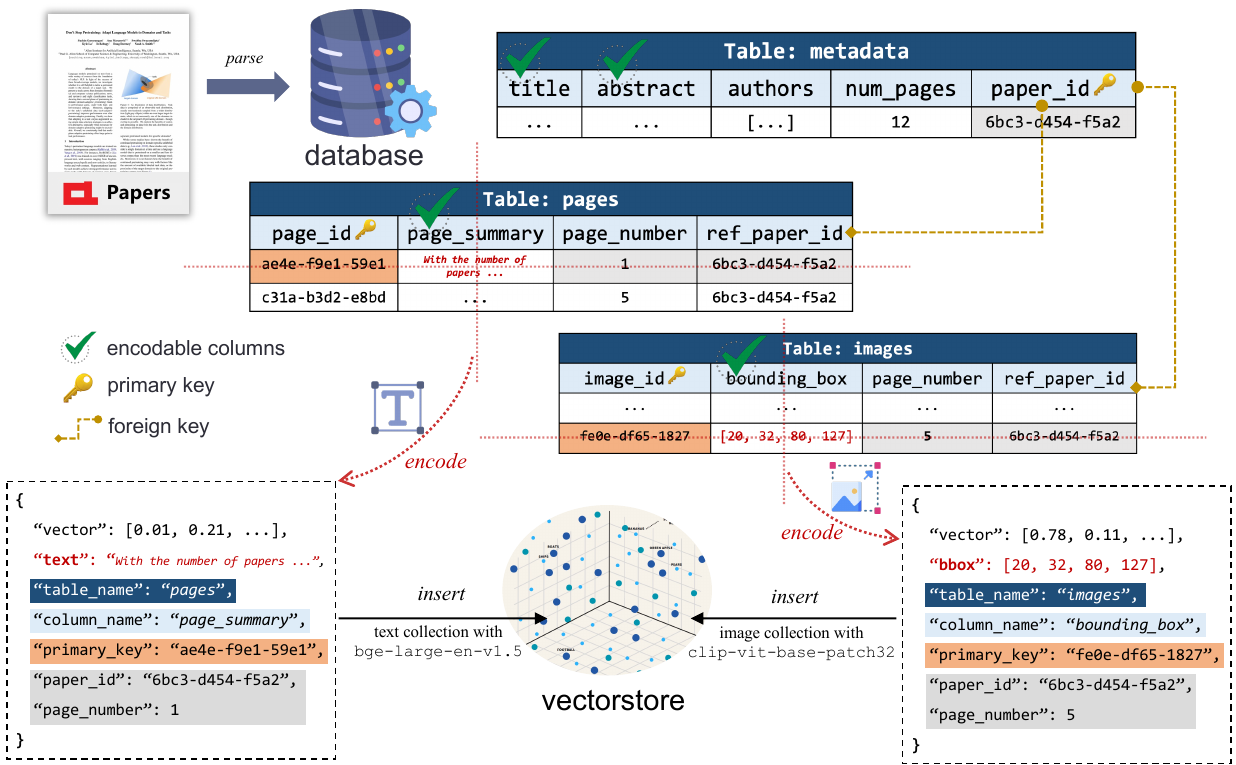}
    \caption{Illustration of how to convert an encodable cell value in the database to one data entry in the vectorstore.}
    \label{fig:db2vs}
\end{figure}

After the completion of the symbolic part, we switch to the neural encoding of parsed elements. Firstly, we label each column in the DB schema as ``\underline{encodable}'' or not. For text modality, we mark those columns of data type ``{\tt varchar}'' as encodable whose cell values are relatively long~(e.g., column ``{\tt pages.page\_summary}'' in Figure~\ref{fig:db2vs}); while for images, the ``{\tt bounding\_box}'' columns which record $4$ coordinates of figures or tables are recognized to extract the square region in the PDF. Next, we resort to encoding models of both modalities to vectorize text snippets or cropped images.

To build a one-to-one mapping between each cell value in the DB and neural vector in the VS, we supplement each data point in the VS with its corresponding table name, column name, and primary key value for that row in the DB. This triplet can uniquely identify each value in the DB. Besides, we add $2$ extra fields, namely ``{\tt paper\_id}'' and ``{\tt page\_number}'', into the JSON dict to enable metadata filtering. These data entries will be inserted into the VS, categorized into different collections based on the encoding model and modality~\footnote{We create $4$ collections in the Milvus~\cite{milvus} vectorstore: for text embeddings, we use BM25~\cite{bm25}, {\tt all-MiniLM-L6-v2}~\cite{minilm}, and {\tt bge-large-en-v1.5}~\cite{bge_embedding}, while {\tt clip-vit-base-patch32}~\cite{clip} for images.}.

Through the first two stages~(\cref{sec:pdf_parsing} and \cref{sec:multimoda-encoding}), various chunking perspectives in the VS are intrinsically connected via a structured DB schema~(visualized in Figure~\ref{fig:database_schema_encodable}). Moreover, long-form texts or visual bounding boxes in the DB are vectorized into the VS to support fuzzy semantic matching.

\subsection{Iterative Agent Interaction}
Now that the database and vectorstore have been populated, we can build an RAG agent which proactively retrieves context from both the DB and VS.
\subsubsection{Action Space Design}
\label{sec:actions}
Firstly, we introduce the $5$ parameterized actions with arguments that agents can take during interaction, namely \retrievefromvectorstore, \retrievefromdatabase, \viewimage, \calculateexpr, and \generateanswer.
\paragraph{\retrievefromvectorstore} This action converts the classic static retrieval into real-time dynamic retrieval. It has multiple arguments which are adjustable~(see Lst.~\ref{lst:retrievefromvectorstore}), e.g.,
``{\tt query}'' encourages LLMs to rewrite the user intention more clearly~\cite{query-rewrite}, while
``{\tt table\_name}'' and ``{\tt column\_name}'' request the agent to select an appropriate perspective for retrieval. During execution, the environment will retrieve context concerning ``{\tt query}'' with specified constraints ``{\tt filter}''~(e.g., {\tt page\_number=1}) from the VS and return the observation in a tabular format~(see App.~\ref{app:observation_format}).
\begin{lstlisting}[language=Python, caption={\retrievefromvectorstore action with its parameters in function calling format.},label={lst:retrievefromvectorstore}]
RetrieveFromVectorstore(
    # user input can be rephrased
    query: str,
    # select encoding model/modality
    collection_name: str, 
    # (table_name, column_name) together defines which view to search
    table_name: str,
    column_name: str,
    # allow fine-grained meta filtering
    filter: str = '',
    limit: int = 5
)
\end{lstlisting}

\paragraph{\retrievefromdatabase} This action accepts a single parameter ``{\tt sql}''. It operates by executing the provided SQL query against the prepared DB and fetching the resulting table to the agent. Similar to ToolSQL~\cite{tool-sql}, through iterative symbolic retrieval, agents can predict different SQLs to explore the DB and exploit multiple pre-parsed views on interpreting PDF content.

\paragraph{\viewimage} To address potential PDF parsing errors during the first stage~(\cref{sec:pdf_parsing}) and leverage features of more advanced VLMs, we devise this action to extract a cropped region~(defined by ``{\tt bounding\_box}'') in one PDF page and send the {\tt base64} encoded image back to agents. Notably, the concrete coordinates can be obtained through retrieval from either the DB or VS during interaction. This way, the agent will acquire the desired image as observation and then reason based on it.
\begin{lstlisting}[language=Python, caption={\viewimage action with its parameters in function calling format.},label={lst:viewimage}]
ViewImage(
    paper_id: str,
    page_number: int,
    # 4-tuple of float numbers, if [], return the image of entire page
    bounding_box: List[float] = []
)
\end{lstlisting}

\paragraph{\calculateexpr} Inspired by \citet{calculator}, we also include this action which accepts a Python ``{\tt expr}'' (e.g., \texttt{2 + 3 * 4}) and returns the calculation result. This simple integration stably reduces hallucination \emph{w.r.t.} math problems in our pilot study.

\paragraph{\generateanswer} This is the \emph{terminal} action that returns the final answer when agents determine that the retrieved context suffices to resolve the input question. The only parameter ``{\tt answer}'' can be of any type depending on the user requirement. Refer to App.~\ref{app:action_definition} for formal action definitions.

\subsubsection{Hybrid Neural Symbolic Retrieval}
In each turn, agents predict one action~(see App. \ref{app:action_format} for different action formats) to interact with the environment and obtain the real-time observation. We adopt the popular ReAct~\cite{react} framework, which requires agents to output their thought process first~(see the right part of Figure~\ref{fig:framework}).
This iterative retrieval~\cite{iter-retgen} enables agents to leverage complementary strengths of both retrieval paradigms in continuous steps.

For example, agents can firstly filter relevant rows by executing a SQL program in the database. \retrievefromdatabase is particularly adept at handling structured queries and metadata-based constraints. After extracting primary key values of rows for the intermediate output,
we can further narrow down the final result set
by inserting those key values
into the ``{\tt filter}'' parameter of action \retrievefromvectorstore to conduct neural semantic matching, i.e., 
\begin{center}
\small
{\tt filter=\textquotesingle primary\_key in [pk values of sql exec]\textquotesingle}.
\end{center}
Conversely, agents can firstly query the VS via neural search to select the most relevant entries, and then treat this transitional set as a temporary table or condition in the subsequent SQL retrieval.
\section{Experiment}
\begin{figure*}[htbp]
    \centering
    \includegraphics[width=0.95\textwidth]{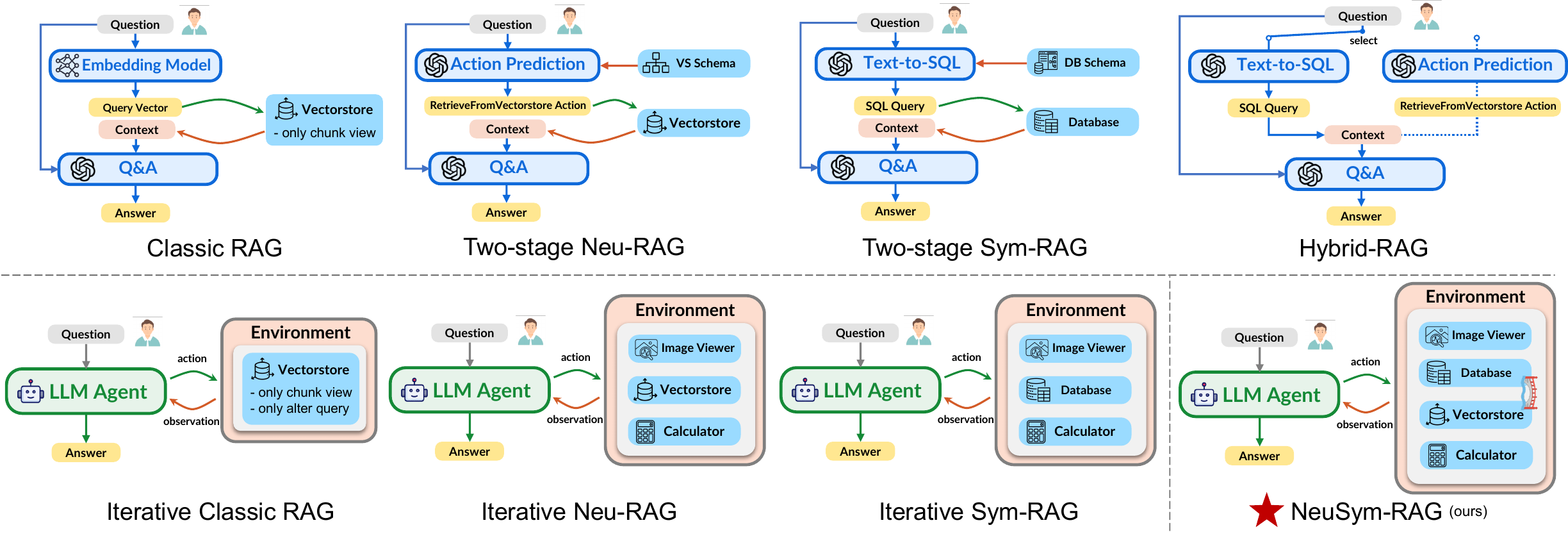}
    \caption{Comparison of different structured baselines with our \ours. Refer to App.~\ref{app:agent_baseline} for text description.
    }
    \label{fig:baseline}
\end{figure*}
This section introduces our human-labeled dataset \dataset, the main experiment and ablation study. Code and data are publicly available \footnote{\url{https://github.com/X-LANCE/NeuSym-RAG}}.

\begin{table}[htbp]
\centering
\resizebox{0.48\textwidth}{!}{
\begin{tabular}{c|m{19em}}
    \hline

    \hline
    \textbf{Category} & \textbf{Question} \\
    \hline \hline
    text & What are the main components of ERRA model? \\
    \hline
    table & On the dataset proposed by this work, how much does the GPT-3.5-turbo model improve its GPT4score after using Graph-CoT? \\
    \hline
    image & Considering the performance of ChatDev agent on DSEval-LeetCode benchmark, what is the most common cause of the errors?  \\
    \hline
    formula & How does Multi-DYLE combine the three different losses as the objective of training? \\
    \hline
    metadata & Which conference was the paper \textquotesingle Fact-Checking Complex Claims with Program-Guided Reasoning\textquotesingle\ published in? Is it a long, short or findings? \\
    \hline

    \hline
\end{tabular}
}
\caption{Examples of single-doc type from \dataset dataset. See App.\ref{app:extra_samples} for the other two types. 
}
\label{tab:dataset_examples}
\end{table}

\subsection{Q\&A Dataset on AI Research Papers}
\label{sec:dataset}
\paragraph{\dataset} Previous Q\&A datasets on academic papers typically focus on simple questions based on a single page or merely the abstract of the PDF, which is far from reflecting real-world scenarios. To address this, we annotate a more complex Q\&A dataset regarding full and even multiple PDFs, called \dataset. Based on published AI papers in $2023$ and $2024$,
\numannotator researchers 
manually annotate \numquestion questions which span across: a) $3$ task types (i.e., single-doc details, multi-doc analysis, and paper retrieval), b) $5$ categories (i.e., text, table, image, formula, and metadata), and c) $2$ evaluation genres (i.e., hard-coding objective metrics and LLM-based subjective assessment). We showcase one example for each category in Table~\ref{tab:dataset_examples}.
\paragraph{Other Benchmarks} To validate the universality, we also convert two other full-PDF-based Q\&A datasets, namely M3SciQA \cite{m3sciqa} and SciDQA \cite{scidqa}. These two benchmarks have $452$ and $2937$ test samples, respectively. And we adjust them into the data format of \dataset. See App.~\ref{app:dataset_details} for more details about datasets.

\paragraph{Evaluation Metrics}
While most long-form Q\&A datasets utilize LLMs or human experts for accuracy (including the aforementioned M3SciQA and SciDQA), we propose more flexible and precise evaluation metrics. Specifically, we 1) adopt answer formatting similar to DABench~\cite{infiagent}, which imposes Python-style restrictions on output (e.g., ``\emph{Your answer should be a Python list of two strings.}''), and 2) implement instance-specific, execution-based evaluation by designing \metricsize functions with optional parameters. These functions are categorized as either subjective or objective, depending on whether they involve the judgement from LLMs (see App.~\ref{sec:evaluation_metrics_detail}).

\begin{table*}[htbp]
  \centering
  
  \resizebox{0.99\textwidth}{!}{
  \begin{tabular}{l|cccccc|ccc|cccc}
    \hline

    \hline
    \multirow{2}[2]{*}{\textbf{Model}} & \multicolumn{6}{c|}{\textbf{\dataset}}           & \multicolumn{3}{c|}{\textbf{M3SciQA}} & \multicolumn{4}{c}{\textbf{SciDQA}} \\
          & \textbf{text} & \textbf{table} & \textbf{image} & \textbf{formula} & \textbf{metadata} & \textbf{AVG} & \textbf{table} & \textbf{image} & \textbf{AVG} & \multicolumn{1}{l}{\textbf{table}} & \multicolumn{1}{l}{\textbf{image}} & \multicolumn{1}{l}{\textbf{formula}} & \multicolumn{1}{l}{\textbf{AVG}} \\
    \hline\hline
    \multicolumn{14}{c}{\textbf{Classic-RAG}} \\
    \hline
    GPT-4o-mini & 12.3  & 11.9  & 12.5  & 16.7  & 13.6  & 13.4  & 17.9  & 10.6  & 15.6  & 59.4  & 60.4  & 59.3  & 59.8 \\
    GPT-4V & 13.2  & 13.9  & 10.0  & 13.9  & 13.6  & 14.7  & 12.1  & 8.8   & 11.1  & 56.6  & 56.8  & 58.1  & 57.4 \\
    Llama-3.3-70B-Instruct & 8.7   & 7.9   & 9.5   & 16.7  & 0.0   & 10.0  & 12.7  & 8.1   & 11.3  & 56.8  & 58.8  & 58.9  & 58.0 \\
    Qwen2.5-VL-72B-Instruct & 9.6   & 5.9   & 11.9  & 11.1  & 13.6  & 10.5  & 11.6  & 11.6  & 11.6  & 54.8  & 56.9  & 56.3  & 56.2 \\
    DeepSeek-R1 & 11.7  & 13.9  & 9.5   & 30.6  & 9.1   & 13.9  & 11.9  & 9.5   & 11.2  & 63.9  & 61.3  & 61.7  & 62.4 \\
    \hline\hline
    \multicolumn{14}{c}{\textbf{\ours}} \\
    \hline
    GPT-4o-mini & 33.0  & 12.9  & 11.9  & 19.4  & 18.2  & 30.7  & 18.7  & 16.6  & 18.0  & 63.0    & 63.6  & 62.5  & 63.0 \\
    GPT-4V & 38.9  & \textbf{18.8} & \textbf{23.8}  & \textbf{38.9}  & \textbf{27.3} & 37.3 & 13.7 & 13.4 & 13.6 & 62.6 & 63.5 & 63.2 & 63.1 \\
    Llama-3.3-70B-Instruct & 30.6 & 11.9 & 16.7 & 16.7 & \textbf{27.3} & 29.3  & \textbf{26.3} & 17.6 & \textbf{23.6} & 55.5 & 57.3 & 56.6 & 56.4 \\
    Qwen2.5-VL-72B-Instruct & \textbf{43.4}  & 15.8  & 11.9  & 25.0  & \textbf{27.3}  & \textbf{39.6}  & 20.2 & \textbf{22.7} &	21.1 & 60.2  & 60.6  & 61.8  & 60.5 \\
    DeepSeek-R1 & 33.2  & 16.8  & 11.9  & 27.8  & 18.2  & 32.4  & 19.0  & 13.7  & 17.4  & \textbf{64.3}  & \textbf{64.6}  & \textbf{63.9}  & \textbf{64.5} \\
    \hline

    \hline
  \end{tabular}
  }
  \caption{Main results of different LLMs using Classic-RAG or \ours on three datasets.}
  \label{tab:main_experiment}
\end{table*}
\begin{table*}[htbp]
\centering
\resizebox{0.99\textwidth}{!}{
\begin{tabular}{l|c|c|c|c|ccccc|c}
    \hline

    \hline
    \textbf{Method} & \textbf{Neural} & \textbf{Symbolic} & \textbf{Multi-view} & \textbf{\# Interaction(s)} & \textbf{sgl.} & \textbf{multi.} & \textbf{retr.} & \textbf{subj.} & \textbf{obj.} & \textbf{AVG} \\
    \hline\hline
    Question only & \multirow{3}{*}{\textcolor[RGB]{176,23,31}{\fontsize{8pt}{8pt}\selectfont\XSolidBrush}} & \multirow{3}{*}{\textcolor[RGB]{176,23,31}{\fontsize{8pt}{8pt}\selectfont\XSolidBrush}} & \multirow{3}{*}{\textcolor[RGB]{176,23,31}{\fontsize{8pt}{8pt}\selectfont\XSolidBrush}} & 1 & 5.7 & 8.0 & 0.4 & 9.4 & 2.7 & 4.0 \\
    Title + Abstract & ~ & ~ & ~ & 1 & 5.7 & 14.0 & 0.0 & 13.1 & 3.6 & 5.4 \\
    Full-text w/. cutoff & ~ & ~ & ~ & 1 & 28.3 & 10.7 & 0.4 & 26.2 & 7.6 & 11.2 \\
    \hline
    Classic RAG & \multirow{2}{*}{\textcolor[RGB]{48,128,20}{\fontsize{8pt}{8pt}\selectfont\Checkmark}} & \multirow{2}{*}{\textcolor[RGB]{176,23,31}{\fontsize{8pt}{8pt}\selectfont\XSolidBrush}} & \multirow{2}{*}{\textcolor[RGB]{176,23,31}{\fontsize{8pt}{8pt}\selectfont\XSolidBrush}} & 1 & 18.2 & 4.0 & 9.4 & 8.4 & 11.0 & 10.5 \\
    Iterative Classic RAG & ~ & ~ & ~ & $\ge$ 2 & 8.2 &	10.0 & 15.2  & 5.6 & 13.2 & 11.8 \\
    \hline
    Two-stage Neu-RAG & \multirow{2}{*}{\textcolor[RGB]{48,128,20}{\fontsize{8pt}{8pt}\selectfont\Checkmark}} & \multirow{2}{*}{\textcolor[RGB]{176,23,31}{\fontsize{8pt}{8pt}\selectfont\XSolidBrush}} & \multirow{2}{*}{\textcolor[RGB]{48,128,20}{\fontsize{8pt}{8pt}\selectfont\Checkmark}} & 2 & 19.5 & 10.0 & 5.3 & 15.9 & 9.4 & 10.7 \\
    Iterative Neu-RAG & ~ & ~ & ~ & $\ge$ 2 & \textbf{37.7} & 18.7 & 48.4 & \textbf{32.7} & 38.3 & 37.3 \\
    \hline
    Two-stage Sym-RAG & \multirow{2}{*}{\textcolor[RGB]{176,23,31}{\fontsize{8pt}{8pt}\selectfont\XSolidBrush}} & \multirow{2}{*}{\textcolor[RGB]{48,128,20}{\fontsize{8pt}{8pt}\selectfont\Checkmark}} & \multirow{2}{*}{\textcolor[RGB]{48,128,20}{\fontsize{8pt}{8pt}\selectfont\Checkmark}} & 2 & 12.2 & 5.4 & 9.4 & 10.6 & 8.7 & 9.1 \\
    Iterative Sym-RAG & ~ & ~ & ~ & $\ge$ 2 & 32.1 & 14.7 & 33.6 & 27.1 & 28.3 & 28.0 \\
    \hline
    Graph-RAG  & \textcolor[RGB]{48,128,20}{\fontsize{8pt}{8pt}\selectfont\Checkmark} & \textcolor[RGB]{176,23,31}{\fontsize{8pt}{8pt}\selectfont\XSolidBrush} & \textcolor[RGB]{48,128,20}{\fontsize{8pt}{8pt}\selectfont\Checkmark} & $2$ & 22.2 & 11.1 & 0.0 & 21.1 & 11.5 & 15.6  \\
    \hline
    Hybrid-RAG  & \multirow{2}{*}{\textcolor[RGB]{48,128,20}{\fontsize{8pt}{8pt}\selectfont\Checkmark}} & \multirow{2}{*}{\textcolor[RGB]{48,128,20}{\fontsize{8pt}{8pt}\selectfont\Checkmark}} & \multirow{2}{*}{\textcolor[RGB]{48,128,20}{\fontsize{8pt}{8pt}\selectfont\Checkmark}} & $2$ & 23.3 & 9.3 & 5.7 & 16.8 & 10.5 & 11.8 \\
    \textbf{\ours (ours)} & ~ & ~ & ~ & $\ge$ 2 & 28.3 & \textbf{32.3} & \textbf{58.2} & 27.1 & \textbf{42.6} & \textbf{39.6} \\
    \hline

    \hline
\end{tabular}
}
\caption{Performances of different RAG methods on \dataset dataset, where ``\# Interaction(s)'' denotes the number of LLM calls for a single question. See App.~\ref{app:agent_baseline} for detailed introduction of each RAG baseline.}
\label{tab:method_ablation}
\end{table*}

\subsection{Experiment Settings}
\label{sec:exp_set}
\paragraph{Baselines} We implement $11$ methods (partly illustrated in Figure \ref{fig:baseline}), ranging from trivial input questions, to more advanced structured models. The main differences lie in: a) Single-view or Multi-view: the Classic RAG only has access to text chunks, while the others can select from multiple pre-parsed perspectives; b) Two-stage or Iterative: the major distinction between the two rows in Figure \ref{fig:baseline} is whether agentic retrieval is permitted; c) Composition of the environment: the complete backend contains relational database, vectorstore, image viewer and calculator. Different baselines may only include part of them, which further restricts the permissible actions~(see App.~\ref{app:action_definition}).


\paragraph{LLMs and Hyper-parameters} We evaluate \ours with various LLMs. For closed-source ones, we use GPT-4o-mini-2024-07-18 and GPT-4-1106-vision-preview. Regarding state-of-the-art open-source LLMs, we include InternLM2.5 \cite{InternLM2}, Llama-3.3-70B-Instruct \cite{llama3.3}, Qwen2.5-VL-72B-Instruct \cite{Qwen2.5-VL}, and DeepSeek-R1 \cite{deepseekr1}. As for hyper-parameters, the $\mathrm{temperature}$ is set to $0.7$ and $\mathrm{top\_p}$ is fixed to $0.95$. The maximum retrieved tokens in each turn and the cutoff for full-text input are both limited to $5k$. And the threshold of interaction turns is $20$. Detailed resource consumption, including running time and LLM API cost, is presented in App.~\ref{app:supply}.


\subsection{Main Experiment}

Based on our pilot study~(Table~\ref{tab:format_ablation} in App.~\ref{app:supply}), we fix the action format to \texttt{markdown} and the observation format to \texttt{json} for all experiments below.

In Table~\ref{tab:main_experiment}, we show performances of our proposed method on different LLMs and datasets. Accordingly,
1) \ours remarkably outperforms the Classic RAG baseline across all datasets, with a minimal \gap improvement on \dataset for all LLMs. With more customizable actions and more trials, \ours can interact with the backend environment with higher tolerance and learn through observations to retrieve proper information for question answering.
2) VLMs perform better in tasks that require vision capability, e.g., in M3SciQA where LLMs have to view an anchor image in the first place.
3) Open-source LLMs are capable of handling this complicated interactive procedure in a zero-shot paradigm, and even better than closed-source LLMs. \ours with Qwen2.5-VL-72B-Instruct elevates the overall accuracy by $29.1\%$, compared to the Classic RAG baseline. Surprisingly, it surpasses GPT-4o-mini by an impressive $8.9\%$, which highlights the universality of \ours.

Next, to figure out the contribution of each component in \ours, we compare the performances of different structured RAG agent methods described in \cref{sec:exp_set}. From Table~\ref{tab:method_ablation}, we can observe that:
1) Two-stage Neu-RAG outperforms Classic RAG, while Hybrid RAG achieves even further improvements.
This consistent increase can be attributed to the fact that agents can adaptively determine the parameters of actions, e.g., which chunking view to select for neural retrieval.
2) Iterative retrievals are superior to their two-stage variants. Through multi-turn interaction, the agent can explore the backend environment and select the most relevant information to answer the question.
3) As the number of interactions increases, objective scores rise faster than subjective scores, indicating that with more retrievals, LLMs generate more rationally. On the whole, \ours defeats all adversaries. It verifies that providing multiple views and combining two retrieval strategies both contribute to the eventual performance.

\subsection{Ablation Study}
For brevity, in this section, \texttt{GPT} denotes GPT-4o-mini and \texttt{Qwen} represents Qwen2.5-72B-Instruct. And unless otherwise specified, we utilize \texttt{Qwen}.

\begin{figure}[htbp]
    \centering
    \includegraphics[width=0.45\textwidth]{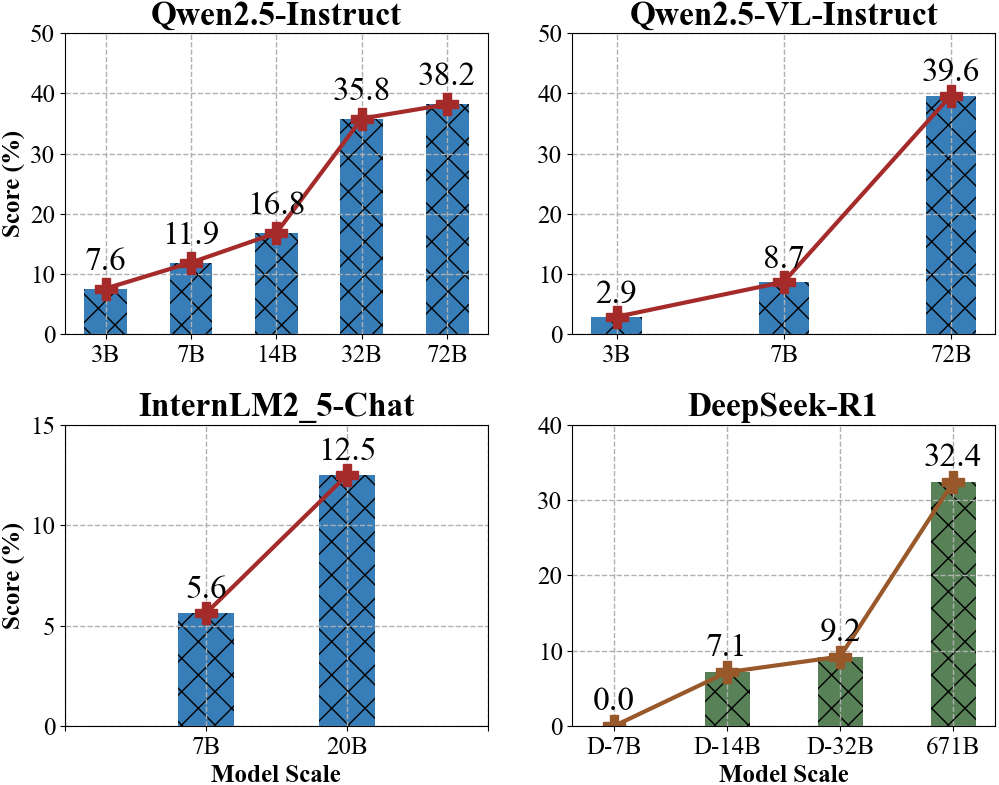}
    \caption{Performances of different model sizes on \dataset dataset. ``D-'' denotes ``Distill-Qwen-''.}
    \label{fig:scale_abl}
\end{figure}

\paragraph{Different Model Scales}
In Figure~\ref{fig:scale_abl}, we show the performance changes by varying the model scales. Accordingly, we observe that: 1) The performances consistently improve as the scales of LLMs increase. 2) Unfortunately, the R1-Distill-Qwen series models lag far behind their counterparts (Qwen2.5-Instruct). Although they generate valuable thoughts during the interaction, they struggle to exactly follow the action format specified in our prompt. We hypothesize that they severely overfit the instructions during R1 post-training.

\paragraph{Chunking with Multiple Views}
In the Classic RAG baseline, we segment raw documents following continuous $512$ tokens. We wonder whether different chunking views lead to varied outcomes. Results in Table~\ref{tab:chunking_ablation} are insightful: the classic chunking strategy is still the best in general. However, chunking views tailored to specific aspects may achieve the best results in their particular categories.
\begin{table}[htbp]
\centering
\resizebox{0.48\textwidth}{!}{
\begin{tabular}{c|cccc|c}
    \hline

    \hline
    \textbf{Retrieved} & \multirow{2}{*}{\textbf{text}} & \multirow{2}{*}{\textbf{table}} & \multirow{2}{*}{\textbf{image}} & \multirow{2}{*}{\textbf{formula}} & \textbf{Overall} \\
    \textbf{Column} &  &  &  &  & \textbf{(\%)}\\
    \hline\hline
    \texttt{chunk\_content} & 13.95 & \textbf{15.00} & \textbf{22.22} & 28.57 & \textbf{18} \\
    \hline
    \texttt{section\_content} & \textbf{18.60} & 10.00 & 5.56 & 14.29 & 12 \\
    \hline
    \texttt{table\_content} & 13.95 & \textbf{15.00} & 0.00 & 9.52 & 9 \\
    \hline
    \texttt{image\_summary} & 4.65 & 0.00 & 16.67 & 4.76 & 6 \\
    \hline
    \texttt{equation\_content} & 6.98 & 0.00 & 5.56 & \textbf{33.33} & 8 \\
    \hline

    \hline
\end{tabular}
}
\caption{Performances of different retrieval options of the Classic RAG baseline on a subset of \dataset.}
\label{tab:chunking_ablation}
\end{table}

\begin{figure*}[htbp]
    \centering
    \includegraphics[width=1.0\textwidth]{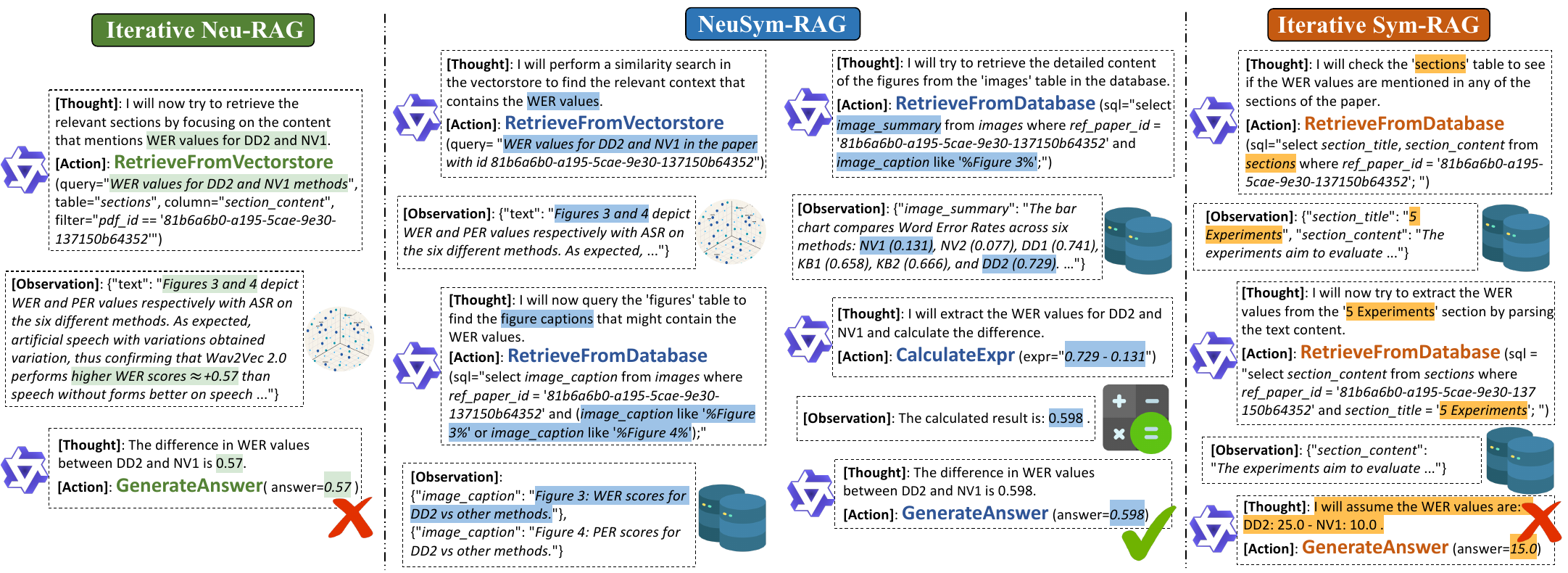}
    \caption{Case study on \dataset dataset. The input question is ``\emph{In terms of WER values with ASR across the six different methods tested in the paper, how much higher is DD2 compared to NV1?}''.}
    \label{fig:case_study}
\end{figure*}
\begin{table}[htbp]
\centering
\resizebox{0.49\textwidth}{!}{
\begin{tabular}{c|c|c|ccccc|c}
    \hline

    \hline
    \multirow{2}{*}{\textbf{BM25}} & \multirow{2}{*}{\textbf{\texttt{MiniLM}}} & \multirow{2}{*}{\textbf{\texttt{bge}}} & \multirow{2}{*}{\textbf{text}} & \multirow{2}{*}{\textbf{table}} & \multirow{2}{*}{\textbf{image}} & \multirow{2}{*}{\textbf{formula}} & \multirow{2}{*}{\textbf{meta}} & \textbf{Overall} \\
     &  & &  &  &  & &  & \textbf{(\%)} \\
    \hline \hline
    \textcolor[RGB]{48,128,20}{\fontsize{8pt}{8pt}\selectfont\Checkmark} & \textcolor[RGB]{176,23,31}{\fontsize{8pt}{8pt}\selectfont\XSolidBrush} & \textcolor[RGB]{176,23,31}{\fontsize{8pt}{8pt}\selectfont\XSolidBrush} & 48.8 & 35.0 & 38.9 & 61.9 & 45.5 & 46 \\
    \textcolor[RGB]{176,23,31}{\fontsize{8pt}{8pt}\selectfont\XSolidBrush} & \textcolor[RGB]{48,128,20}{\fontsize{8pt}{8pt}\selectfont\Checkmark} & \textcolor[RGB]{176,23,31}{\fontsize{8pt}{8pt}\selectfont\XSolidBrush} & 41.9 & 25.0 & 27.8 & 33.3 & 27.3 & 33  \\
    \textcolor[RGB]{176,23,31}{\fontsize{8pt}{8pt}\selectfont\XSolidBrush} & \textcolor[RGB]{176,23,31}{\fontsize{8pt}{8pt}\selectfont\XSolidBrush} & \textcolor[RGB]{48,128,20}{\fontsize{8pt}{8pt}\selectfont\Checkmark} & 39.5 & 25.0 & 16.7 & 52.4 & 54.6 & 37 \\
    \textcolor[RGB]{48,128,20}{\fontsize{8pt}{8pt}\selectfont\Checkmark} & \textcolor[RGB]{48,128,20}{\fontsize{8pt}{8pt}\selectfont\Checkmark} & \textcolor[RGB]{176,23,31}{\fontsize{8pt}{8pt}\selectfont\XSolidBrush} & 44.2 & 30.0 & 27.8 & 52.4 & 27.3 & 40 \\
    \textcolor[RGB]{48,128,20}{\fontsize{8pt}{8pt}\selectfont\Checkmark} & \textcolor[RGB]{176,23,31}{\fontsize{8pt}{8pt}\selectfont\XSolidBrush} & \textcolor[RGB]{48,128,20}{\fontsize{8pt}{8pt}\selectfont\Checkmark} & 41.9 & 20.0 & 33.3 & 42.9 & 27.4 & 35 \\
    \textcolor[RGB]{176,23,31}{\fontsize{8pt}{8pt}\selectfont\XSolidBrush} & \textcolor[RGB]{48,128,20}{\fontsize{8pt}{8pt}\selectfont\Checkmark} & \textcolor[RGB]{48,128,20}{\fontsize{8pt}{8pt}\selectfont\Checkmark} & 34.9 & 10.0 & 16.7 & 47.6 & 45.5 & 30 \\
    \textcolor[RGB]{48,128,20}{\fontsize{8pt}{8pt}\selectfont\Checkmark} & \textcolor[RGB]{48,128,20}{\fontsize{8pt}{8pt}\selectfont\Checkmark} & \textcolor[RGB]{48,128,20}{\fontsize{8pt}{8pt}\selectfont\Checkmark} & 39.5 & 20.0 & 16.7 & 47.6 & 45.5 & 35 \\
    \hline

    \hline
\end{tabular}
}
\caption{Performance of \ours on \dataset dataset with different text encoding model~(s).}
\label{tab:encodings}
\end{table}
\paragraph{Choices of Text Embedding Models} Table~\ref{tab:encodings} demonstrates the performance of \ours with different vector collections~(\emph{i.e.}, encoding models). The experimental results uncover that a single BM25 text collection suffices to attain the best performance. We speculate that there might be two possible reasons for this observation: 1) existing LLMs are still not capable of exploring retrieved context from multiple encoding models and then selecting the optimal one. This behavioral pattern requires profound reasoning abilities. 
2) Another potential insight is that, for different encoding models, the best practice is to directly choose the highest-performing one instead of leaving the burden to the agent. And in our preliminary experiments, we also find similar phenomena that instead of creating different DB tables to store the parsed PDF content with different auxiliary tools, directly choosing the top-performing tools, that is PyMuPDF~\cite{pymupdf} and MinerU~\cite{mineru} in \cref{sec:pdf_parsing}, and using a single DB table achieves the best performance.

\begin{figure}[htbp]
    \centering
    \includegraphics[width=0.45\textwidth]{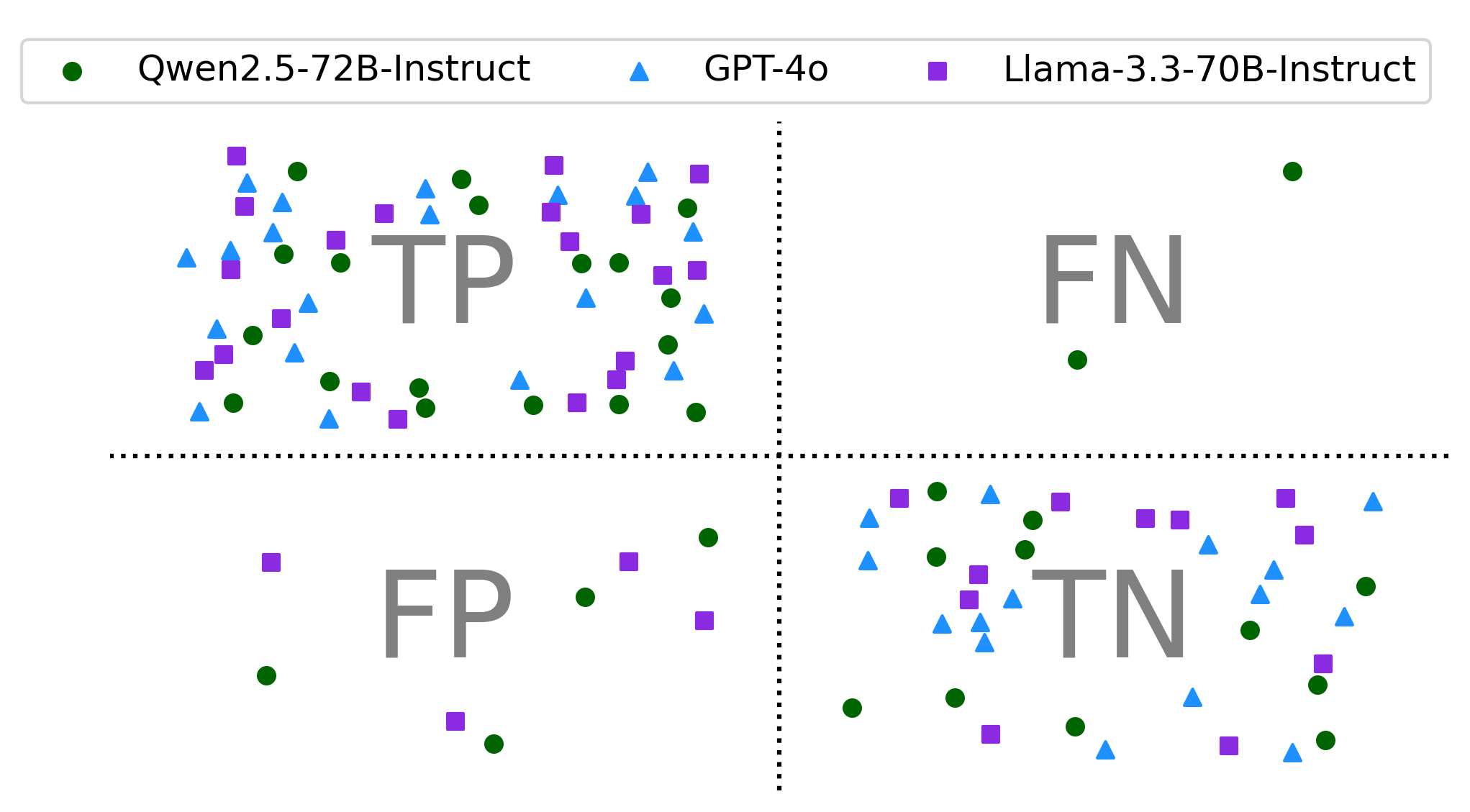}
    \caption{The consistency between human and LLM-based evaluation. We classify examples that are judged as correct by human as positive samples.}
    \label{fig:eval_abl}
\end{figure}

\begin{table}[htbp]
\centering
\resizebox{0.48\textwidth}{!}{
\begin{tabular}{c|c|c|c|c}
    \hline

    \hline
    \textbf{Agent} & \textbf{Summary} & \multirow{2}{*}{\textbf{objective}} & \multirow{2}{*}{\textbf{subjective}} & \textbf{Overall} \\
    \textbf{Model} & \textbf{Model} &  &  & \textbf{(\%)} \\
    \hline\hline
    \multirow{2}{*}{\texttt{GPT}} & \texttt{GPT} & 24.62 & 34.29 & 28 \\
    & \texttt{Qwen} & 23.08 & 34.29 & 27 \\
    \hline
    \multirow{2}{*}{\texttt{Qwen}} & \texttt{GPT} & 30.77  & 57.14 & 40 \\
    & \texttt{Qwen} & 30.77 & 54.28 & 39 \\
    \hline

    \hline
\end{tabular}
}
\caption{Performances of different models with different LLM summary models on a subset of \dataset.} 
\label{tab:process_ablation}
\end{table}
\paragraph{Details of Pre-processing and Evaluation}
Aiming to reduce the high cost of API calls, we explore the impact of using open-source v.s. closed-source LLMs during 1) PDF pre-parsing (the summary generation), and 2) subjective evaluation. As shown in Table~\ref{tab:process_ablation}, cheaper open-source LLMs appear to be a viable alternative to closed-source ones, since the final results are insensitive to element summaries produced by different LLMs.
As for evaluation, we randomly sample $100$ long-form answers requiring subjective assessment and recruit AI researchers to independently measure the semantic equivalence compared to the reference answer.
Figure~\ref{fig:eval_abl} indicates that, in most cases, LLM evaluations align closely with human judgments. Notably, closed-source GPT-4o aligns exactly with human judgments, whereas open-source LLMs are occasionally either too strict or too lenient. Therefore, we choose closed-source LLMs for subjective evaluation throughout the experiments.

\subsection{Case Study}




In Figure~\ref{fig:case_study}, we serialize the interaction trajectory of $3$ iterative methods for the same question.
\ours firstly predicts a \retrievefromvectorstore action to search chunks regarding ``\emph{WER values}''. After detecting relevant figures, it predicts a \retrievefromdatabase action to access their captions and summaries. In contrast, Iterative Neu-RAG fails to exploit the figures due to the absence of the DB, and Iterative Sym-RAG struggles to locate relevant context.
In this case, we can clearly observe one benefit or mode of combining both retrieval actions: the VS helps agents identify relevant context, while the DB for precise queries. This synergy achieves the accurate answer unattainable by either retrieval type alone, emphasizing the significance of their collaboration.

\section{Related Work}
\label{sec:related_work}

\paragraph{Structured and Iterative Retrieval} Previous literature has proposed various RAG methods for structured knowledge sources. GraphRAG \cite{graphrag} builds a knowledge graph~(KG) to summarize partial and final answers. HybridRAG \cite{hybridrag} integrates KG-based and vector-based methods for financial documents. FastRAG \cite{fastrag} employs schema and script learning to extract structured data, combining text search with KG.
Another promising line of work proves that agents can benefit from iterative retrieval. IRCoT \cite{ircot} handles multi-step QA by interleaving retrieval with chain-of-thought. Iter-RetGen \cite{iter-retgen} integrates retrieval and generation iteratively for collective knowledge.
CoA \cite{chain-of-agents} uses multiple agents for long-context tasks, with a manager agent combining their results.
Our \ours integrates neural and symbolic retrieval in an agentic interaction to fetch multi-view context.

\paragraph{PDF-based Q\&A Benchmarks} Existing question answering datasets over PDF documents often overlook the underlying structure and layout. QASPER \cite{qasper} generate questions from merely titles and abstracts, while FinQA \cite{finqa} targets single-page Q\&A. QASA \cite{qasa} retains section structures but misses other key elements like figures and tables. More recent works, like Visconde \cite{visconde}, M3SciQA \cite{m3sciqa}, and SciDQA \cite{scidqa}, focus on long-form subjective evaluation. 
This work addresses these limitations by tackling questions from full PDFs of AI research papers, including objective metrics.
\section{Conclusion}
In this work, we propose a hybrid neural symbolic retrieval framework (\ours) for semi-structured PDF-based question-answering~(Q\&A).
Experiments on three human-labeled realistic datasets regarding AI research, especially on the self-annotated \dataset, show that \ours remarkably outperforms the Classic RAG baseline by more than \gap points. Future works include: 1) training a specialized LLM agent of medium size to further improve performances and efficiency,
and 2) extending Q\&A tasks to other vertical domains such as healthcare and law which heavily rely on external well-structured PDF files.

\section*{Limitations}
Despite the superiority of the proposed \ours framework, it still has the following limitations:
1) The pipeline of parsing and encoding the PDF content is relatively slow, especially when higher parsing accuracy is demanded. That is, we need OCR models typically with larger capacity and longer processing time. This method is suitable if all documents are pre-stored in a repository, since the pre-processing can be conducted once and for all. However, it 
may be unsatisfactory
in scenarios where users expect real-time uploads of new PDFs.
2) The iterative chain-of-retrieval procedure incurs extra time delay compared to classic RAG. This is anticipated and \citet{snell2024scaling} points out that scaling inference computation is more effective than scaling model parameters.
3) Different from GraphRAG~\cite{graphrag} which can pre-parse any free-form texts, \ours requires that the external documents exhibit implicit structures or layouts. And PDF files in AI research naturally conform to this constraint. Thus, integrating the two retrieval paradigms can maximize the utilization of the inherent advantages of semi-structured documents.
4) Currently, \dataset
are limited to 
the domain of AI research papers.
We are actively working on
expanding our efforts to other vertical domains, including finance and law.

\section*{Ethics Statement}
All papers utilized in the construction of our curated Q\&A dataset, \dataset, are sourced from the websites {\tt ACL Anthology}~\footnote{ACL Anthology: \url{https://aclanthology.org/}}, {\tt arXiv.org}~\footnote{arXiv.org API: \url{https://info.arxiv.org/help/api/index.html}}, and {\tt OpenReview.net}~\footnote{OpenReview.net APIv2: \url{https://docs.openreview.net/reference/api-v2}}, which are all licensed under the Creative Commons Attribution 4.0 International License (CC BY 4.0).
The use of these materials complies fully with their licensing terms, strictly for academic and research purposes.
No private, sensitive, or personally identifiable information is used in this work. The study adheres to the ethical guidelines outlined by ACL, ensuring integrity, transparency, and reproducibility.

\section*{Acknowledgements}
We would like to thank Hankun Wang, Danyang Zhang, Situo Zhang, Yijie Luo, Ye Wang, Zichen Zhu, Dingye Liu, and Zhihan Li for their careful annotation on the \dataset dataset. We also thank all reviewers for their constructive and thoughtful advice. This work is funded by the China NSFC Projects (62120106006, 92370206, and U23B2057) and Shanghai Municipal Science and Technology Major Project (2021SHZDZX0102).

\bibliography{anthology,custom}
\bibliographystyle{acl_natbib}

\appendix
\onecolumn
\clearpage
\section{\dataset Dataset}
\label{app:dataset_details}
\subsection{Data Format}
\label{app:example_format}
In this section, we briefly introduce the data format of our \dataset dataset. Each data instance is represented as a JSON dictionary which contains the following fields:

\begin{itemize}[leftmargin=15pt]
    \item {\tt uuid}: Globally unique uuid of the current task example.
    \item {\tt question}: The user question about the given papers.
    \item {\tt answer\_format}: The requirements for the output format of LLMs, e.g., a Python list of text strings or a single float number, such that we can evaluate the answer conveniently.
    \item {\tt tags}: A list of tags denoting different types, categories or genres. Feasible tags include [``single'', ``multiple'', ``retrieval'', ``text'', ``image'', ``table'', ``formula'', ``metadata'', ``subjective'', ``objective''].
    \item {\tt anchor\_pdf}: A list of PDF uuids that are directly related to or explicitly mentioned in the question.
    \item {\tt reference\_pdf}: A list of PDF uuids that may or may not help answer the question.
    \item {\tt conference}: A list of conference names plus year. This field is only useful in the ``\emph{paper retrieval}'' setting to limit the scope of search space, as shown in the task example of Listing~\ref{lst:airqa_example_retrieval}.
    \item {\tt evaluator}: A dictionary containing 2 fields, {\tt eval\_func} and {\tt eval\_kwargs}, which defines how to evaluate the model outputs. Concretely,
    \begin{itemize}[label=\tiny\(\blacksquare\)]
        \item the ``{\tt eval\_func}'' field defines the name of our customized Python function (or metric) which is used to compare the predicted result and the expected ground truth;
        \item the ``{\tt eval\_kwargs}'' field defines the arguments for the corresponding evaluation function, which usually contain the gold or reference answer and other optional parameters.
    \end{itemize}
    For example, in Listing~\ref{lst:airqa_example_multiple}, we utilize the function ``{\tt eval\_structured\_object\_exact\_match}'' to evaluate the LLMs output according to the given gold answer [``SCG-NLI'', ``false''].
\end{itemize}

\begin{lstlisting}[language=json, caption={A multi-doc analysis task example of JSON format from the dataset \dataset.}, label={lst:airqa_example_multiple}]
{
    "uuid": "927ff9af-42f7-5216-a6f9-f106e8ff6759",
    "question": "On the HEML sentence level with AUC metric, which baseline outperforms MIND on specific conditons? Is it the best variant according to the paper that proposed that baseline?",
    "answer_format": "Your answer should be a Python list of two strings. The first string is the name of the baseline (with variant) that outperforms MIND, as proposed in the anchor PDF. The second string is either `true` or `false`.",
    "tags": [
        "multiple",
        "text",
        "table",
        "objective"
    ],
    "anchor_pdf": [
        "621d42a1-dbab-5003-b7c5-625335653001"
    ],
    "reference_pdf": [
        "ab661558-432d-5e5e-b49c-a3660a40986e",
        "1a21b653-3db0-55e8-9d34-8b6cd3dcbefa",
        "85111b8b-4df0-5a9a-8d11-a7ae12eebcf6",
        "0597ce2b-cd8c-5b5b-b692-e8042d8548de",
        "6df0f3f3-e2e1-5d7a-9d70-3114ceac5939",
        "02f7fff5-cec7-5ac8-a037-f5eb117b9547"
    ],
    "conference": [],
    "evaluator": {
        "eval_func": "eval_structured_object_exact_match",
        "eval_kwargs": {
            "gold": [
                "SCG-NLI", 
                "false"
            ],
            "ignore_order": false,
            "lowercase": true
        }
    }
}
\end{lstlisting}

\begin{lstlisting}[language=json, caption={A paper retrieval task example of JSON format from the dataset \dataset.}, label={lst:airqa_example_retrieval}]
{
    "uuid": "4b4877cd-4cdc-5d52-ac20-edfaa6dd7e32",
    "question": "Is there any paper leverages knowledge distillation of language models for textual out-of-distribution detection or anomaly detection?",
    "answer_format": "Your answer should be the title of the paper WITHOUT ANY EXPLANATION.",
    "tags": [
        "retrieval",
        "text",
        "objective"
    ],
    "anchor_pdf": [],
    "reference_pdf": [],
    "conference": [
        "acl2023"
    ],
    "evaluator": {
        "eval_func": "eval_paper_relevance_with_reference_answer",
        "eval_kwargs": {
            "reference_answer": "Multi-Level Knowledge Distillation for Out-of-Distribution Detection in Text"
        }
    }
}
\end{lstlisting}

\subsection{Data Examples}
\label{app:extra_samples}
Table~\ref{tab:dataset_examples_app} shows $3$ examples in \dataset with different task types, namely single-doc details, multi-doc analysis, and paper retrieval. Different numbers of PDF documents are involved in different task types. For instance, single-doc examples only specify one paper in its ``{\tt anchor\_pdf}'' field, while multi-doc examples must mention multiple papers in the ``{\tt anchor\_pdf}'' or ``{\tt reference\_pdf}'' fields. And for paper retrieval, we relax the search scope to the entire conference venue, such as ``{\tt acl2023}'' in Listing~\ref{lst:airqa_example_retrieval}.
\begin{table*}[htbp]
\centering
\resizebox{0.99\textwidth}{!}{
\begin{tabular}{c|m{21em}|m{19em}}
    \hline

    \hline
    \textbf{Category} & \textbf{Question} & \textbf{Answer Format} \\
    \hline

    \hline
    {\tt single} & On the ALFWorld dataset experiments, how much did the success rate improve when the authors used their method compared to the original baseline model? & Your answer should be a floating-point number with one decimal place. \\
    \hline
    {\tt multiple} & I would like to reproduce the experiments of KnowGPT, could you please provide me with the websites of the datasets applied in the experiment? & Your answer should be a Python list of 3 strings, the websites. Note that you should provide the original URL as given in the papers that proposed the datasets. \\
    \hline
    {\tt retrieval} & Find the NLP paper that focuses on dialogue generation and introduces advancements in the augmentation of one-to-many or one-to-one dialogue data by conducting augmentation within the semantic space. & Your answer should be the title of the paper WITHOUT ANY EXPLANATION. \\
    \hline

    \hline
\end{tabular}
}
\caption{Demonstration examples of different task types from the dataset \dataset, where {\tt single}, {\tt multiple} and {\tt retrieval} represent single-doc details, multi-doc analysis and paper retrieval, respectively.}
\label{tab:dataset_examples_app}
\end{table*}

\subsection{Dataset Statistics}

\begin{figure}[htb]
    \centering
    \begin{minipage}[c]{0.5\textwidth}
        \centering
	\begin{tabular}{c|c|c}
        \hline
        
        \hline
        \textbf{Data Splits} & \textbf{Number} & \textbf{Ratio(\%)} \\
        \hline\hline
        single & 159 & 29 \\
        multiple & 150 & 27 \\
        retrieval & 244 & 44 \\
        \hline
        text & 470 & 85 \\   
        table & 101 & 18 \\
        image & 42 & 8 \\
        formula & 36 & 7 \\
        metadata & 22 & 4 \\
        \hline
        objective & 446 & 81 \\
        subjective & 107 & 19\\
        \hline
        \textbf{Overall} & 553 & 100 \\
        \hline
        
        \hline
        \end{tabular}
        \captionof{table}{The task number and ratio of different data splits on the dataset \dataset.}
        \label{tab:airqa_100_statistics}
    \end{minipage}
    \begin{minipage}[c]{0.45\textwidth}
        \centering
	\includegraphics[width=0.8\textwidth]{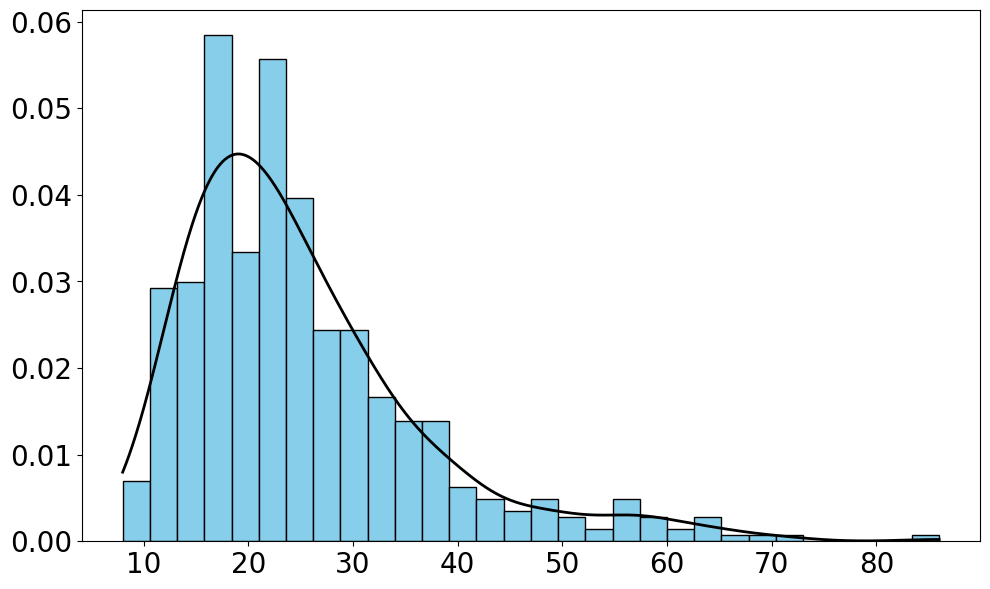}
		\caption{Frequency distribution of question lengths.}
        \label{fig:question_length}
        \includegraphics[width=0.55\textwidth]{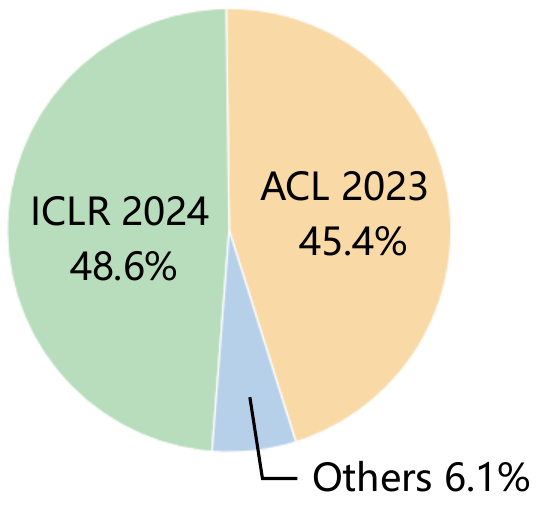}
        \caption{Conference distribution of the papers used.}
        \label{fig:conferences}
    \end{minipage}
\end{figure}

In total, the entire \dataset contains \numpdf PDF documents and \numquestion questions with respect to AI research papers mostly in year 2023 and 2024. Among them, $6384$ PDFs are published papers from ACL 2023 and ICLR 2024, while the PDFs left are also from publicly available websites \texttt{arXiv.org} and \texttt{OpenReview.net}. Table~\ref{tab:airqa_100_statistics}, Figure~\ref{fig:question_length} and Figure~\ref{fig:conferences} show the statistics of the dataset \dataset. Note that, one task example can be labeled with more than one category tag (i.e., text, table, image, formula, and metadata). As for the \numquestion examples, $240$ are converted from the LitSearch~\cite{litsearch} benchmark (a paper retrieval task set). And only examples in LitSearch whose ground truth belongs to ACL 2023 or ICLR 2024 venues are selected and revised to be compatible with our \dataset data format. Correspondingly, we set the value of field ``\texttt{conference}'' (defined in App.~\ref{app:example_format}) to be ``{\tt acl2023}'' or ``{\tt iclr2024}'' in order to restrict the paper search space.

\subsection{Other Full-PDF-based Academic Research Q\&A Benchmarks}
\label{app:other_benchmarks}
Apart from \dataset, we also convert two other existing Q\&A benchmarks M3SciQA~\cite{m3sciqa} and SciDQA~\cite{scidqa} in academic research to validate the universality of our \ours framework, which gives us $492$ and $2937$ more test samples, respectively. Specifically,
\begin{description}
    \item[M3SciQA] The answers of the test set on M3SciQA are not publicly available yet till \today. Thus, we only utilize the complete $452$ labeled samples in the validation set as the test suite which are licensed under Creative Commons Attribution 4.0. Notice that, each question in M3SciQA involves multiple papers. To obtain the final answer, agents need to firstly find the correct paper which belongs to the references of the anchor paper. Since our constructed database contains the entire paper set on M3SciQA, we only provide the title of the anchor PDF without referenced paper titles in the prompt to make the testing scenario more realistic. And agents need to find the target paper title by itself in the reference section. PDFs of all relevant papers can be downloaded from the arXiv.org website. As for evaluation, we follow the official evaluation metric~\footnote{Evaluation for M3SciQA: \url{https://github.com/yale-nlp/M3SciQA/blob/main/src/evaluate_detail.py}} and leverage large language model based assessment with exactly the same prompt, model~(gpt-4-0125-preview) and temperature~($0.0$) to determine whether the final long-form answer is correct.
    \item[SciDQA] We convert the original $2937$ test samples which are licensed under Open Data Commons Attribution License (ODC-By) v1.0. The relevant paper PDFs can all be downloaded from OpenReview.net and we categorize each input question as ``table'', ``image'' and ``formula'' based on heuristic rules (e.g., whether a specific keyword such as ``equation'' is mentioned in the question or ground truth). As for evaluation, we adopt the official LLM-based judgement and answer extraction scripts~\footnote{Evaluation for SciDQA: \url{https://github.com/yale-nlp/SciDQA/tree/main/src/evals}} except that the large language model is replaced with gpt-4o-mini from Llama-3.1-70B-Instruct due to its performance being more closely aligned with humans according to Figure~\ref{fig:eval_abl}.
\end{description}
Two separate evaluation functions are encapsulated for these two benchmarks (see Table~\ref{tab:evaluation_metrics_detail}). The authors declare that the usage of these two existing benchmarks strictly obeys the aforementioned licenses.
\section{Supplementary Experiments and Settings}
\label{app:supply}
For \ours, we also investigate different formats of the action and observation spaces~(exemplified in App.~\ref{app:action_format} and App.~\ref{app:observation_format}). According to Table~\ref{tab:format_ablation}: 1) The action format has a more substantial impact on the results compared to the observation format. 2) Specifically, both LLMs perform the best when the action format is ``\emph{markdown}'', which resembles the function calling fashion in Python. Another possible explanation is that the other $3$ action formats generally impose stricter formatting requirements. 3) Different LLMs may prefer different observation formats on account of their training corpus. But the gaps, especially for the top-performing $2$ choices, are relatively small.
\begin{table}[htbp]
\centering
\resizebox{0.8\textwidth}{!}{
\begin{tabular}{c|c|c|c|c|c}
    \hline

    \hline
    \textbf{Type} & \textbf{Format} & \textbf{Overall(\%)} & \textbf{Type} & \textbf{Format} & \textbf{Overall(\%)}\\
    \hline\hline
    \multicolumn{3}{c|}{GPT-4o-mini-2024-07-18} & \multicolumn{3}{c}{Qwen2.5-72B-Instruct} \\
    \hline
    \multirow{4}[0]{*}{action} & \textbf{markdown} & \textbf{40} & \multirow{4}[0]{*}{action} & \textbf{markdown} & \textbf{28} \\
    ~ & json & 35 & ~ & json & 27 \\
    ~ & xml & 28 & ~ & xml & 19 \\
    ~ & yaml & 32 & ~ & yaml & 26 \\
    \hline
    \multirow{4}[0]{*}{observation} & \textbf{json} & \textbf{40} & \multirow{4}[0]{*}{observation} & json  & 28 \\
    ~ & markdown & 31 & ~ & markdown & 27 \\
    ~ & html & 31 & ~ & html & 26 \\
    ~ & string & 39 & ~ & \textbf{string} & \textbf{30} \\
    \hline

    \hline
\end{tabular}
}
\caption{Ablation study on different action and observation formats on a subset of \dataset dataset.}
\label{tab:format_ablation}
\end{table}

In Table~\ref{tab:supplementary_statistics}, we present some statistics of iterative methods on different models in our experiments. In general, GPT-series LLMs complete the task with fewer interaction rounds, indicating that closed-source LLMs exhibit stronger capabilities and confidence. Using the Qwen2.5-72B-Instruct LLM, the \ours approach completes the task with more interaction rounds than the other two iterative methods, since the \ours approach can integrate the context from both retrieval paradigms.

\begin{table*}[htbp]
\centering
\resizebox{0.99\textwidth}{!}{
\begin{tabular}{c|c|ccccc}
    \hline

    \hline
    \textbf{Model} & \textbf{RAG Method} & \textbf{\# Interaction(s)} & \textbf{\# Prompt Token(s)} & \textbf{\# Completion Token(s)} & \textbf{Time~(s)} & \textbf{Cost~(\$)} \\
    \hline\hline
    Llama-3.3-70B-Instruct & \ours & 4.45 & 36521 & 832 & 40 & - \\
    \hline
    \multirow{3}{*}{Qwen2.5-72B-Instruct} & Iterative Sym-RAG & 5.16 & 43548 & 2730 & 105 & - \\
    ~ & Iterative Neu-RAG & 4.19 & 48600 & 2395 & 94 & - \\
    ~ & \ours & 5.26 & 58262 & 1987 & 83 & - \\
    \hline
    GPT-4o-mini & \ours & 3.59 & 29306 & 518 & 20 & 0.0059 \\
    \hline
    GPT-4V & \ours & 3.61 & 54657 & 1098 & 17 & 0.1464 \\
    \hline

    \hline
\end{tabular}
}
\caption{Statistics of the number of interaction(s), accumulated prompt / completion token(s), time consumption, and LLM cost per sample with different models and RAG methods on $100$ samples from \dataset.}
\label{tab:supplementary_statistics}
\end{table*}

\section{Agent Baselines}
\label{app:agent_baseline}
In this section, we elaborate the implementation of each RAG method utilized in Table~\ref{tab:method_ablation}.

\begin{itemize}[leftmargin=*]
\item \textbf{Classic RAG}
fetches question-related chunks from the vectorstore and directly provides them as the context for LLMs to answer the question. The chunk size is fixed to $512$ tokens using the \texttt{RecursiveCharacterTextSplitter} from \texttt{langchain}~\footnote{Langchain text splitter: \url{https://python.langchain.com/docs/how_to/recursive_text_splitter/}.} and the retrieved top-$K$ size is set to $4$. In other words, we pre-fetch the content of column ``{\tt chunks.text\_content}'' based on the raw input question and insert them as the context in the prompt. The default text embedding model is fixed to the widely used sentence transformer {\tt all-MiniLM-L6-v2}~\cite{minilm}.

\item \textbf{Iterative Classic RAG} enables LLMs to repeatedly alter their query texts and iteratively retrieve chunks until the answer can be obtained. But the view is always fixed to column ``{\tt chunks.text\_content}''. It can be regarded as one simple implementation of the popular IRCoT~\cite{ircot}.

\item \textbf{Two-stage Neu-RAG} splits the task into two stages. At stage one,
LLMs predict a \retrievefromvectorstore action with only one chance. But they can predict the parameters in that action, e.g., the query text, the (table, column) perspective to choose, the embedding collection, and the returned top-$K$ size. While at stage two, agents must output the final answer with retrieved context.

\item \textbf{Iterative Neu-RAG} is developed from the two-stage one, incorporating additional chances in multi-turn interactions. LLMs can predict multiple parameterized actions to retrieve from the vectorstore until the interaction trajectory suffices to answer the question. And we incorporate another two useful actions \viewimage and \calculateexpr (formally defined in \cref{sec:actions} and App.~\ref{app:action_definition}) into the multi-turn interaction. 

\item \textbf{Two-stage Sym-RAG} requires LLMs to generate a SQL query first.
After SQL execution upon the backend database, they must predict the answer using the retrieved context, with only one chance. Indeed, this symbolic retrieval belongs to another field of text-to-SQL~\cite{lgesql}.

\item \textbf{Iterative Sym-RAG} is the multi-round version, where LLMs can iteratively interact with the symbolic database and try different SQLs to better conclude the final answer. The additional two action types \viewimage and \calculateexpr are also included in the action space like Iterative Neu-RAG.

\item \textbf{Graph-RAG}~\cite{graphrag} is implemented by following the official guideline of library \texttt{graphrag}~\footnote{GraphRAG website: \url{https://microsoft.github.io/graphrag/get_started/}.}. And we adopt the \texttt{local} search mode since it performs better than \texttt{global} in our pilot study. To reduce the graph construction time, we build a separate graph for each PDF document, and restrict the retrieval scope for each question to its tied papers.

\item \textbf{Hybrid-RAG}~\cite{hybridrag} can be considered as a two-stage structured baseline, where the LLM agent firstly determines which action (\retrievefromvectorstore or \retrievefromdatabase) to use. After predicting the parameters of the action, and fetching the context from either the vectorstore or database, it needs to generate the final answer with only one chance.
\end{itemize}

\section{Evaluation Metrics}
\label{sec:evaluation_metrics_detail}
    The detailed information of all $18$ evaluation functions we design is presented in Table~\ref{tab:evaluation_metrics_detail}.
\sloppy
\begin{table*}[htbp]
\centering
\resizebox{0.99\textwidth}{!}{
\begin{tabular}{c|c|m{10.5em}|m{19.5em}}
    \hline

    \hline
    \textbf{Eval Type} & \textbf{Sub-Type} & \textbf{Function} & \textbf{Description} \\
    \hline \hline
    \multirow{18}[0]{*}{\textit{objective}} & \multirow{10}[0]{*}{match} & eval\_bool\_exact\_match & Evaluate the output against the answer using exact boolean match. \\
    \cline{3-4}
    ~ & ~ & eval\_float\_exact\_match & Evaluate the output against the answer using exact float match with variable precision or tolerance. \\
    \cline{3-4}
    ~ & ~ & eval\_int\_exact\_match & Evaluate the output against the answer using exact integer match. \\
    \cline{3-4}
    ~ & ~ & eval\_string\_exact\_match & Evaluate the output against the answer using exact string match. \\
    \cline{3-4}
    ~ & ~ & eval\_structured\_object
        \_exact\_match & Evaluate the output against the answer recursively by parsing them both as Python-style lists or dictionaries. \\
    \cline{2-4}
    ~ & \multirow{5}[0]{*}{set} & eval\_element\_included & Evaluate whether the output is included in the answer list. \\
    \cline{3-4}
    ~ & ~ & eval\_element\_list\_included & Evaluate whether each element in the output list is included in the answer list. \\
    \cline{3-4}
    ~ & ~ & eval\_element\_list\_overlap & Evaluate whether the output list overlaps with the answer list. \\
    \cline{2-4}
    ~ & retrieval & eval\_paper\_relevance\_with
        \_reference\_answer & Evaluate whether the retrieved paper is the same as the reference answer. \\
    \hline \hline
    \multirow{9}[0]{*}{\textit{subjective}} & \multirow{5}[0]{*}{semantic} & eval\_reference\_answer
        \_with\_llm & Evaluate the output against the reference answer using LLMs. \\
    \cline{3-4}
    ~ & ~ & eval\_scoring\_points\_with
        \_llm & Evaluate whether the scoring points are all mentioned in the output using LLMs. \\
    \cline{3-4}
    ~ & ~ & eval\_partial\_scoring\_points
        \_with\_llm & Evaluate whether the scoring points are partially mentioned in the output using LLMs. \\ 
    \cline{2-4}
    ~ & formula & eval\_complex\_math\_form
        ula\_with\_llm & Evaluate the mathematical equivalence between the output and the answer formatted in Latex using LLMs. \\
    \hline\hline
    \multicolumn{2}{c|}{\multirow{8}{*}{\it logical}} & eval\_conjunction & Evaluate the conjunction of multiple evaluation functions. The output passes the evaluation if and only if all the elements in the output pass the corresponding sub-evaluations. \\
    \cline{3-4}
     \multicolumn{2}{c|}{} & eval\_disjunction & Evaluate the disjunction of multiple evaluation functions. The output passes the evaluation if and only if at least one of the element in the output passes the corresponding sub-evaluation. \\
    \cline{3-4}
     \multicolumn{2}{c|}{}   & eval\_negation & Evaluate the negation of an evaluation function. The output passes the evaluation if and only if it doesn't pass the original evaluation function. \\
    \hline\hline
    \multicolumn{2}{c|}{\multirow{3}{*}{\it others}} & eval\_m3sciqa & Evaluate examples in dataset M3SciQA with the encapsulated original LLM-based function. \\
    \cline{3-4}
      \multicolumn{2}{c|}{}  & eval\_scidqa & Evaluate examples in dataset SciDQA with the encapsulated original LLM-based function. \\
    \hline

    \hline
\end{tabular}
}
\caption{The checklist of the $18$ used evaluation functions, including their categories, names, and descriptions.}
\label{tab:evaluation_metrics_detail}
\end{table*}

\clearpage
\section{Database Schema and Encodable Columns in Vectorstore}
\label{app:database_schema}

\begin{figure*}[htbp]
    \centering
    \includegraphics[width=0.7\textwidth]{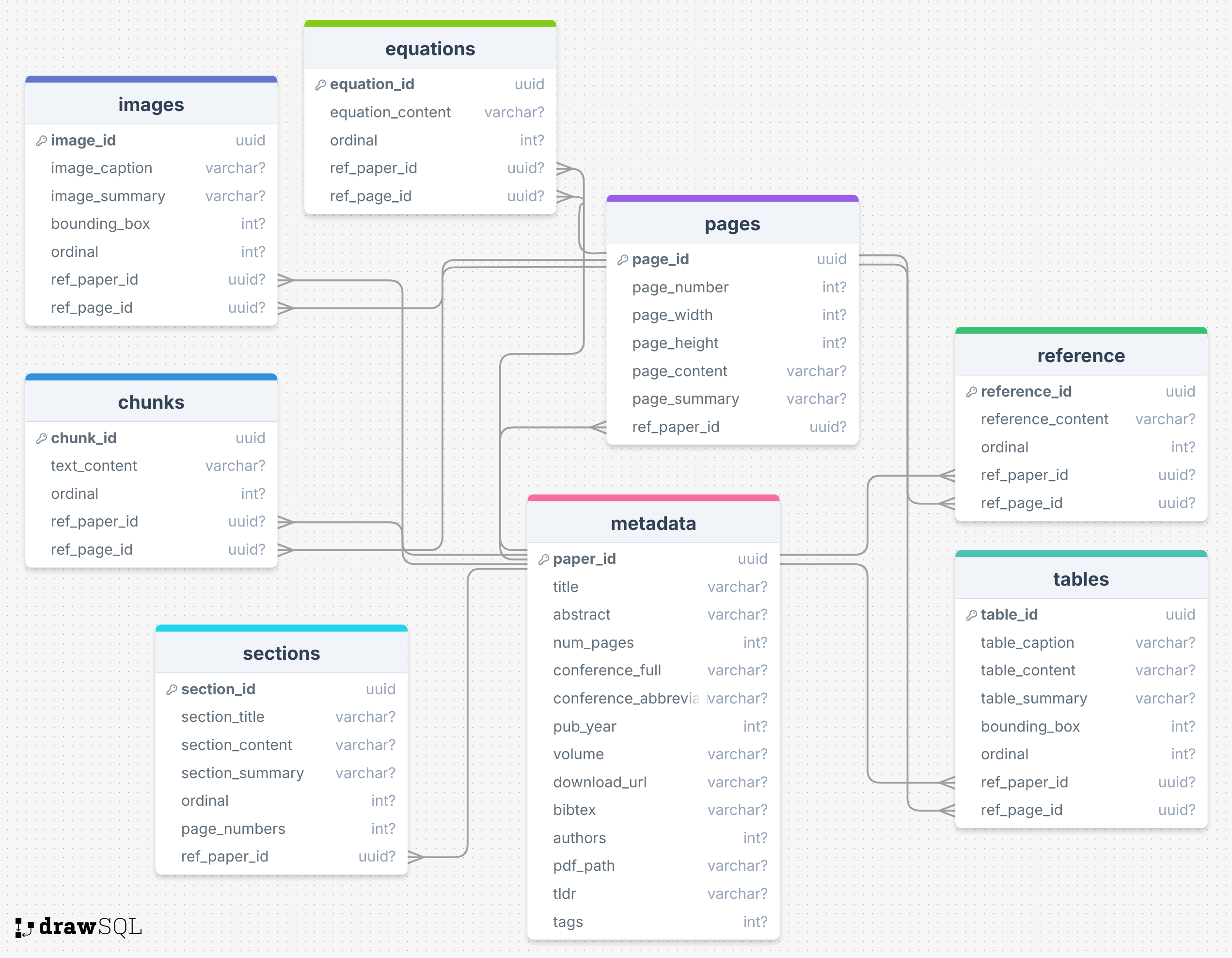}
    \caption{The complete and universal database schema to store the parsed elements of each PDF file. Note that, since the visualization tool {\tt drawSQL} cannot display ``{\tt ARRAY}'' data types, the actual data types of the columns ``\emph{bounding\_box}'', ``\emph{authors}'', ``\emph{tags}'', and ``\emph{page\_numbers}'' are {\tt INT[4]}, {\tt VARCHAR[]}, {\tt VARCHAR[]}, and {\tt INT[]}, respectively.}
    \label{fig:database_schema}
\end{figure*}

\begin{figure*}[htbp]
    \centering
    \includegraphics[width=0.65\textwidth]{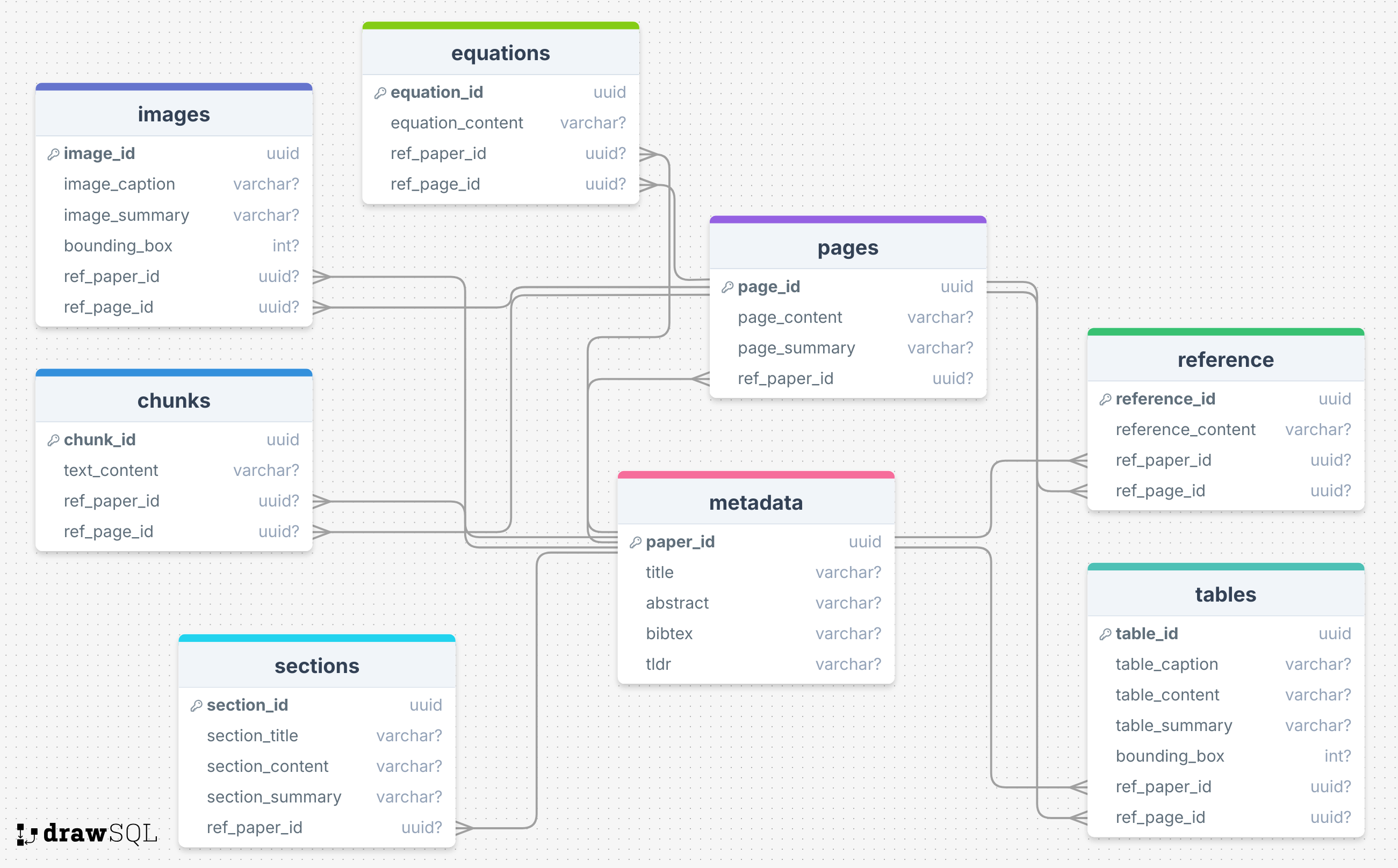}
    \caption{All encodable columns which are inherently connected by the schema sub-graph. Note that, since the visualization tool {\tt drawSQL} cannot display ``{\tt ARRAY}'' data types, the actual data types of columns ``\emph{images.bounding\_box}'' and ``\emph{tables.bounding\_box}'' are both {\tt INT[4]}.}
    \label{fig:database_schema_encodable}
\end{figure*}

The complete database schema to store parsed PDF content is visualized in Figure~\ref{fig:database_schema}, with free online tool drawSQL~\footnote{\url{https://drawsql.app/}}. All encodable columns are listed in Table \ref{tab:schema_and_encodable_columns}. These columns provide various perspectives towards interpreting the PDF content. And these different views are inherently connected by the sub-graph of the original database schema~(as illustrated in Figure~\ref{fig:database_schema_encodable}).

\begin{table*}[!ht]
\renewcommand{\arraystretch}{1.5}
\begin{tabular}{c|c|m{27em}}
\hline

\hline
\textbf{Table}                     & \textbf{Column}             & \textbf{Description}                                                                                                                                                                                                                                                        \\ \hline \hline
\multirow{5}{*}{metadata} & title              & The title of this paper.                                                                                                                                                                                                                                           \\ \cline{2-3} 
                          & abstract           & The abstract of this paper.                                                                                                                                                                                                                                        \\ \cline{2-3} 
                          & bibtex             & The bibtex of this paper.                                                                                                                                                                                                                                          \\ \cline{2-3} 
                          & tldr               & A brief summary of the paper's main idea or findings generated by LLM based on title and abstract.                                                                                                                                                                 \\ \hline \hline
\multirow{2.5}{*}{pages}    & page\_content      & The content of the page.                                                                                                                                                                                                                                           \\ \cline{2-3} 
                          & page\_summary      & A brief summary of the page content, generated by LLM, focusing on key information and describing the page content.                                                                                                                                                \\ \hline \hline
\multirow{6}{*}{images}   & image\_caption     & The caption of this image, empty string if not found.                                                                                                                                                                                                              \\ \cline{2-3} 
                          & image\_summary     & A brief summary of the image, generated by LLM, focusing on key information and describing the image.                                                                                                                                                              \\ \cline{2-3} 
                          & bounding\_box      & The bounding box of the figure in the format $[x_0, y_0, w, h]$, where $(x_0, y_0)$ represents the coordinates of the top-left corner and $(w, h)$ represents the width and height which are used to determine the shape of the rectangle. The cropped image is encoded. \\ \hline \hline
\multirow{6}{*}{tables}   & table\_caption     & Caption of the table, showing key information of the table.                                                                                                                                                                                                        \\ \cline{2-3} 
                          & table\_content     & The content of the table in html format.                                                                                                                                                                                                                           \\ \cline{2-3} 
                          & table\_summary     & A brief summary of the table content generated by LLM, focusing on key information and describing the table content.                                                                                                                                               \\ \cline{2-3} 
                          & bounding\_box      & The bounding box of the table in the format $[x_0, y_0, w, h]$, where $(x_0, y_0)$ represents the coordinates of the top-left corner and $(w, h)$ represents the width and height. The cropped image is encoded.                                                         \\ \hline \hline
\multirow{3.5}{*}{sections} & section\_title     & The title of the current section.                                                                                                                                                                                                                                  \\ \cline{2-3} 
                          & section\_content   & The text content of the current section.                                                                                                                                                                                                                           \\ \cline{2-3} 
                          & section\_summary   & A brief summary of the section content generated by LLM, focusing on key information and describing the section content.                                                                                                                                           \\ \hline \hline
chunks                    & text\_content      & The text content of the current chunk.                                                                                                                                                                                                                             \\ \hline
equations                 & equation\_content  & Content of the equation in latex format.                                                                                                                                                                                                                           \\ \hline
reference                 & reference\_content & Text content of each reference.                                                                                                                                                                                                                                          \\ \hline

\hline
\end{tabular}
\caption{The checklist of all encodable columns with their descriptions.}
\label{tab:schema_and_encodable_columns}
\end{table*}





\clearpage
\section{Prompt Template}
\label{app:prompt_template}
This section presents the detailed structure of the prompts used in various agent baselines~(Figure~\ref{fig:baseline}), outlining the components that shape the agent's interactions and reasoning process. The overall prompt template includes the following five key components.

\definecolor{1}{HTML}{E67E22}
\definecolor{2}{HTML}{967aa1}
\definecolor{3}{HTML}{718355}
\definecolor{4}{HTML}{800f2f}
\definecolor{5}{HTML}{4a4e69}

\definecolor{6}{HTML}{2980B9}
\definecolor{7}{HTML}{C0392B}
\definecolor{8}{HTML}{9B59B6}


\begin{tcolorbox}[title={Overall Prompt Composition}, colback=white,colframe=black,arc=1mm,boxrule=1pt,left=1mm,right=1mm,top=1mm,bottom=1mm]
\small
\textcolor{1}{\textbf{[System Prompt]}: Defines the agent's role and describes the task to tackle, clarifying the context of the task.}\\
\ - - - -\newline 
\textcolor{2}{ \textbf{[Action and Observation Space Prompt]}: Defines the list of all feasible actions that the agent can take, including the action description, observation space, syntax and parameters, and use cases for each type. The detailed specification varies depending on the action format.}\\
\ - - - -\newline 
\textcolor{4}{ \textbf{[Interaction Framework Prompt]}: Outlines the main interaction procedure and template.}\\
\ - - - -\newline 
\textcolor{5}{\textbf{[Hint Prompt]}: Provides hints or suggestions to help the agent refine its planning and reasoning process.}\\
\ - - - -\newline 
\textcolor{3}{ \textbf{[Task Prompt]}: Defines the task input, usually including the input question, required answer format, retrieved context, database schema, vectorstore schema and operators.}
\end{tcolorbox}

\subsection{System Prompt}
The system prompt is used to define the ultimate goal that agents should achieve, the environment (if exists) that agents can interact with and the expected agent behavior to conduct. Notice that, the prompts for different agent baselines differ from each other.

\begin{tcolorbox}[title={System Prompts for Different Agent Baselines}, colback=white,colframe=1,arc=1mm,boxrule=1pt,left=1mm,right=1mm,top=1mm,bottom=1mm]
\small
\color{1}

\textcolor{black}{\textbf{For NeuSym-RAG:}}\newline
You are an intelligent agent with expertise in \textbf{retrieving useful context from both the DuckDB database and the Milvus vectorstore through SQL execution and similarity search} and \textbf{answering user questions}. You will be given a natural language question concerning PDF files, along with the schema of both the database and the vectorstore. Your ultimate goal is to answer the input question with pre-defined answer format. The DuckDB database contains all parsed content of raw PDF files, while the Milvus vectorstore encodes specific column cells from the database as vectors. You can predict executable actions, interact with the hybrid environment (including database and vectorstore) across multiple turns, and retrieve necessary context until you are confident in resolving the question.

\ - - - -\newline \textbf{\#\# Task Description}
 \newline Each input task consists of the following parts:\newline[Question]: A natural language question from the user regarding PDF files, e.g., Is there any ...?\newline[Answer Format]: Specifies the required format of the final answer, e.g., the answer is \"Yes\" or \"No\" without punctuation.\newline[Database Schema]: A detailed serialized schema of the DuckDB database for reference when generating SQL queries. It includes 1) tables, 2) columns and their data types, 3) descriptions for these schema items, and 4) primary key and foreign key constraints.\newline[Vectorstore Schema]: A detailed serialized schema of the Milvus vectorstore for reference when generating executable retrieval actions with specific parameters. It includes 1) collections, 2) fields, 3) encodable (table, column) pairs in the relational database where the vectorized content originates, and 4) grammar for valid filter rules.\newline
 
\textcolor{black}{\textbf{For Classic RAG:}}

You are intelligent agent who is expert in {\textbf{answering user questions}} based on the retrieved context. You will be given a natural language question concerning a PDF file, and your task is to answer the input question with predefined output format using the relevant information.\newline

\textcolor{black}{\textbf{For Two-stage Neu-RAG:}}\newline
\textcolor{black}{\textbf{[Stage 1]}}\quad You are intelligent agent who is expert in \textbf{predicting a well-formed retrieval action} to search useful information to answer the user question. You will be given a natural language question concerning a PDF file and a vectorstore schema which defines all usable collections and fields in them. The vectorized contents in the vectorstore all come from cell values in another relational database which stores the parsed content of the PDF files. And your task is to predict a parametrized retrieval action to find useful information based on vector similarity search. Please refer to the concrete vectorstore schema to produce a valid retrieval action.\newline
\textcolor{black}{\textbf{[Stage 2]}}\quad You are intelligent agent who is expert in \textbf{answering user question} given the retrieved context. You will be given a natural language question concerning a PDF file and the retrieved context. Your task is to predict the final answer based on given question and context. Please refer to the answer format to produce the valid answer.\newline
\end{tcolorbox}
\newpage
\begin{tcolorbox}[title={System Prompt},colback=white,colframe=1,arc=1mm,boxrule=1pt,left=1mm,right=1mm,top=1mm,bottom=1mm]
\color{1}
\small

\textcolor{black}{\textbf{For Iterative Classic RAG and Iterative Neu-RAG:}}\newline You are intelligent agent who is expert in \textbf{retrieving useful context from the vectorstore based on similarity search} and \textbf{answering user questions}. You will be given a natural language question concerning a PDF file and a vectorstore schema of Milvus, and your ultimate task is to answer the input question with pre-defined output format. The Milvus vectorstore encodes various context from the parsed PDF in multi-views. You can predict executable actions, interact with the vectorstore in multiple turns, and retrieve desired context to help you better resolve the question.

\ - - - - 

\textbf{\#\# Task Description}
\newline Each input task consists of the following parts:\newline[Question]: A natural language question from the user regarding PDF files, e.g., Is there any ...?\newline[Answer Format]: Specifies the required format of the final answer, e.g., the answer is ``Yes'' or ``No'' without punctuation.\newline[Vectorstore Schema]: A detailed serialized schema of the Milvus vectorstore for reference when generating executable retrieval actions with specific parameters. It includes 1) collections, 2) fields, 3) encodable (table, column) pairs in the relational database where the vectorized content originates, and 4) grammar for valid filter rules.\newline

\textcolor{black}{\textbf{For Two-stage Sym-RAG:}}\newline
\textcolor{black}{\textbf{[Stage 1]}}\quad You are intelligent agent who is expert in \textbf{writing SQL programs} to retrieve useful information. You will be given a natural language question concerning a PDF file and a database schema which stores the parsed PDF content, and your task is to predict SQL to retrieve content from the database. Please refer to the concrete database schema to produce the valid SQL.\newline
\textcolor{black}{\textbf{[Stage 2]}}\quad \textcolor{black}{the same as method Two-stage Neu-RAG}\newline
 
\textcolor{black}{\textbf{For Iterative Sym-RAG:}}\newline You are intelligent agent who is expert in \textbf{leveraging SQL programs to retrieve useful information} and \textbf{answer user questions}. You will be given a natural language question concerning a PDF file and a database schema of DuckDB which stores the parsed PDF content, and your ultimate task is to answer the input question with predefined output format. You can predict intermediate SQLs, interact with the database in multiple turns, and retrieve desired information to help you better resolve the question.

\ - - - -\newline \textbf{\#\# Task Description}
 \newline Each input task consists of the following parts:\newline[Question]: A natural language question from the user regarding PDF files, e.g., Is there any ...?\newline[Answer Format]: Specifies the required format of the final answer, e.g., the answer is \"Yes\" or \"No\" without punctuation.\newline[Database Schema]: A detailed serialized schema of the DuckDB database for reference when generating SQL queries. It includes 1) tables, 2) columns and their data types, 3) descriptions for these schema items, and 4) primary key and foreign key constraints.\newline

 \textcolor{black}{\textbf{For Hybrid-RAG:}}\newline
\textcolor{black}{\textbf{[Stage 1]}}\quad You are intelligent agent who is expert in predicting a well-formed retrieval action to search useful information to answer the user question. You will be given a natural language question concerning a PDF file, a database schema which stores the parsed PDF content, and a vectorstore schema which defines all usable collections and fields in them. The vectorized contents in the vectorstore all come from cell values in the database. And your task is to predict a parametrized retrieval action to find useful information. Please refer to the concrete schema to produce a valid retrieval action.

\textcolor{black}{\textbf{[Stage 2]}}\quad \textcolor{black}{the same as methods Two-stage Neu-RAG and Two-stage Sym-RAG}
\end{tcolorbox}

\subsection{Action and Observation Space Prompt}
\label{app:act_obs_prompt}
In total, there are $5$ actions~(see \cref{sec:actions}) for the proposed \ours framework. We choose the \retrievefromvectorstore action as an example, and serialize it in JSON format below:

\begin{tcolorbox}[title={Action and Observation Space Prompt for \ours (JSON format)},colback=white,colframe=2,arc=1mm,boxrule=1pt,left=1mm,right=1mm,top=1mm,bottom=1mm]
\color{2}
\small
\textbf{\#\# Action and Observation Space}

All allowable action types include \textbf{[``RetrieveFromVectorstore'', ``RetrieveFromDatabase'', ``CalculateExpr'', ``ViewImage'', ``GenerateAnswer'']}. Here is the detailed specification in \textbf{JSON} format for them:
\newline
\newline
\textbf{\#\#\# Action Type}

RetrieveFromVectorstore
\newline
\newline
\textbf{\#\#\# Description}
\newline
Given a query text, retrieve relevant context from the Milvus vectorstore. Please refer to the schema of different collections and fields for each stored data entry.
\newline
\newline
\textbf{\#\#\# Observation}\newline The observation space is the retrieved top-ranked entries from the Milvus vectorstore based on input parameters. 
\end{tcolorbox}
\newpage
\begin{tcolorbox}[title={Action and Observation Space Prompt for \ours (JSON format) -- continued},colback=white,colframe=2,arc=1mm,boxrule=1pt,left=1mm,right=1mm,top=1mm,bottom=1mm]
\color{2}
\small
\textbf{\#\#\# Syntax and Parameters (JSON Format)}
\begin{verbatim} {
    "action_type": "RetrieveFromVectorstore",
    "parameters": {
        "query": {
            "type": "str",
            "required": true,
            "description": "This query will be encoded and used to search for relevant context.
                        You can rephrase the user question to obtain a more clear and structured requirement."
        },
        "collection_name": {
            "type": "str",
            "required": true,
            "description": "The name of the collection in the Milvus vectorstore to search for
                relevant context. Please ensure the collection does exist in the vectorstore."
        },

        "table_name": {
            "type": "str",
            "required": true,
            "description": "The table name is used to narrow down the search space. And it will
                be added to the filter condition. Please ensure this table has encodable columns."
        },
        "column_name": {
            "type": "str",
            "required": true,
            "description": "The column name is used to narrow down the search space. And it will
                be added to the filter condition. Please ensure it is encodable in `table_name`."
        },
        "filter": {
            "type": "str",
            "required": false,
            "default": "",
            "description": "The filter condition to narrow down the search space. Please refer to
                the syntax of filter rules. By default, it is empty. It is suggested to restrict
                `primary_key`, `pdf_id`, or `page_number` to refine search results."
        },
        "limit": {
            "type": "int",
            "required": false,
            "default": 5,
            "description": "The number of top-ranked context to retrieve. Please ensure that it is a
                positive integer. And extremely large limit values may be truncated."
        }
    }
}
\end{verbatim}

\textbf{\#\#\# Use Cases~(JSON Format)}\newline\newline \textbf{\#\#\#\# Case 1}\newline Search the Mivlus collection \texttt{text\_bm25\_en}, which use BM25 sparse embeddings, with the filter condition ``\texttt{table\_name == 'chunks' and column\_name == 'text\_content' and pdf\_id == '12345678' and page\_number == 1}'' to restrict the content source and return the top $10$ relevant entries.

\textbf{[Action]:}
\begin{verbatim}
{"action_type": "RetrieveFromVectorstore", "parameters": {"query": "Does this paper discuss LLM-based
    agent on its first page?", "collection_name": "text_bm25_en", "table_name": "chunks",
    "column_name": "text_content", "filter": "pdf_id == '12345678' and page_number == 1", "limit": 10}}
\end{verbatim}
\mbox{}\newline
\textbf{\#\#\#\# Case 2}\newline ... more cases in JSON format ...
\newline
\newline
\ - - - -
\newline\newline
... specification for other types of actions, the prompt can be easily inferred ...
\end{tcolorbox}
\subsubsection{Syntax and Parameters for Other Action Types~(JSON Format)}
\label{app:action_definition}
This subsection formally describes the syntax and parameters for the other $4$ action types, including \retrievefromdatabase, \calculateexpr, \viewimage, \generateanswer.

\begin{lstlisting}[language=json, caption={Syntax and Parameters of JSON Format for Other Action Types in \ours}, label={lst:JSON syntax and parameters}]
{
    "action_type": "RetrieveFromDatabase",
    "parameters": {
        "sql": {
            "type": "str",
            "required": true,
            "description": "The concrete DuckDB SQL query to execute and retrieve results."
        }
    }
},
{
    "action_type": "CalculateExpr",
    "parameters": {
        "expr": {
            "type": "str",
            "required": true,
            "description": "The expression to calculate, e.g., '13 * 42'."
        }
    }
},
{
    "action_type": "ViewImage",
    "description": "You can retrieve the visual information of the paper by taking this action. Please provide the paper id, the page number, and the optional bounding box.",
    "observation": "The observation space is the image that you want to view. We will show you the image according to your parameters. The error message will be shown if there is any problem with the image retrieval.",
    "parameters": {
        "paper_id": {
            "type": "str",
            "required": true,
            "description": "The paper id to retrieve the image."
        },
        "page_number": {
            "type": "int",
            "required": true,
            "description": "The page number (starting from 1) of the paper to retrieve the image."
        },
        "bounding_box": {
            "type": "List[float]",
            "required": false,
            "default": [],
            "description": "The bounding box of the image to retrieve. The format is [x_min, y_min, delta_x, delta_y]. The complete PDF page will be retrieved if not provided."
        }
    }
},
{
    "action_type": "GenerateAnswer",
    "parameters": {
        "answer": {
            "type": "Any",
            "required": true,
            "description": "The final answer to the user question. Please adhere to the answer format for the current question."
        }
    }
}
\end{lstlisting}

Note that, except for the Classic RAG baseline, which has no action space, the feasible action types differ among the other baselines described in Figure~\ref{fig:baseline}:
\begin{itemize}
    \item \textbf{Iterative Classic RAG:} it only accepts the \retrievefromvectorstore and \generateanswer actions. For the former one, we further fix the perspective to be ``{\tt chunks.text\_content}'' and the collection to text embedding model {\tt all-MiniLM-L6-v2};
    \item \textbf{Two-stage Neu-RAG:} it only accepts action \retrievefromvectorstore at the first stage;
    \item \textbf{Iterative Neu-RAG:} all available action types during the iterative neural retrieval includes [\retrievefromvectorstore, \calculateexpr, \viewimage, \generateanswer];
    \item \textbf{Two-stage Sym-RAG:} it only accepts action \retrievefromdatabase at the first stage;
    \item \textbf{Iterative Sym-RAG:} all available action types during the iterative symbolic retrieval includes [\retrievefromdatabase, \calculateexpr, \viewimage, \generateanswer];
    \item \textbf{Hybrid-RAG:} it accepts both actions \retrievefromvectorstore and \retrievefromdatabase at the first stage, and no action at the second stage.
\end{itemize}

\subsubsection{Other Action Formats Apart From JSON}
\label{app:action_format}
This subsection introduces other serialized action formats apart from JSON, namely MARKDOWN, XML and YAML.
Take action \retrievefromvectorstore as an example (others can be easily inferred):
\begin{tcolorbox}[title={MARKDOWN Format}, colback=white,colframe=2,arc=1mm,boxrule=1pt,left=1mm,right=1mm,top=1mm,bottom=1mm]
\small
\color{2}
\textbf{\#\#\# Syntax and Parameters~(MARKDOWN Format)}
\begin{Verbatim}
RetrieveFromVectorstore(query: str, collection_name: str, table_name: str, column_name: str, filter: 
str = '', limit: int = 5)
    - query: str, required. The query text will be encoded and used to search for relevant context.
        You can rephrase the user question to obtain a more clear and structured requirement.
    - collection_name: str, required. The name of the collection in the Milvus vectorstore to
        search for relevant context. Please ensure the collection does exist in the vectorstore.
    - table_name: str, required. The table name is used to narrow down the search space. And it
        will be added to the filter condition. Please ensure this table has encodable columns.
    - column_name: str, required. The column name is used to narrow down the search space. And it
        will be added to the filter condition. Please ensure it is encodable in `table_name`.
    - filter: str, optional, default to ''. The filter condition to narrow down the search space.
        Please refer to the syntax of filter rules. By default, it is empty. It is suggested to
        restrict `primary_key`, `pdf_id`, or `page_number` to refine search results.
    - limit: int, optional, default to 5. The number of top-ranked context to retrieve. Please
        ensure that it is a positive integer. And extremely large limit values may be truncated.
\end{Verbatim}
\textbf{\#\#\# Use Cases~(MARKDOWN Format)}\newline\newline \ \textbf{\#\#\#\# Case 1}\newline Search the Mivlus collection \texttt{text\_bm25\_en}, which use BM25 sparse embeddings, with the filter condition ``\texttt{table\_name == 'chunks' and column\_name == 'text\_content' and pdf\_id == '12345678' and page\_number == 1}'' to restrict the content source and return the top $10$ relevant entries.

\textbf{[Action]:}
\begin{verbatim}
RetrieveFromVectorstore(query="Does this paper discuss LLM-based agent on its first page?",
    collection_name='text_bm25_en', table_name='chunks', column_name='text_content', 
    filter="pdf_id == '12345678' and page_number == 1", limit=10)
\end{verbatim}
\mbox{}\newline\textbf{\#\#\#\# Case 2}
\newline
Perform a vector-based similarity search on all cell values from the `\texttt{abstract}` column in the '\texttt{metadata}' table in the database, using the \texttt{MiniLM-L6-v2} setence transformer embeddings. By default, the top $5$ most relevant entries will be returned.

\textbf{[Action]:}
\begin{verbatim}
RetrieveFromVectorstore(query="Is there any work about the topic structured RAG?", collection_name=
    'text_sentence_transformers_all_minilm_l6_v2', table_name='metadata', column_name='abstract')
\end{verbatim}
\mbox{}\newline\textbf{\#\#\#\# Case 3}
\newline
.. more cases in MARKDOWN format ...
\end{tcolorbox}

\begin{tcolorbox}[title={XML Format}, colback=white,colframe=2,arc=1mm,boxrule=1pt,left=1mm,right=1mm,top=1mm,bottom=1mm]
\small
\color{2}
\textbf{\#\#\# Syntax and Parameters~(XML Format)}
\begin{Verbatim}
<action>
    <action_type>RetrieveFromVectorstore</action_type>
    <parameters>
        <query>
            <type>str</type>
            <required>true</required>
            <description>The query text will be encoded and used to search for relevant context. 
            You can rephrase the original user question to obtain a more clear and structured query 
            requirement.</description>
        </query>
        <collection_name>
            <type>str</type>
            <required>true</required>
            <description>The collection name in the Milvus vectorstore to search for relevant 
            context. Please ensure the collection does exist in the vectorstore.</description>
        </collection_name>
        <table_name>
            <type>str</type>
            <required>true</required>
            <description>The table name is used to narrow down the search space. It will be added 
            to the filter condition. Please ensure this table has encodable columns.</description>
        </table_name>
        <column_name>
            <type>str</type>
            <required>true</required>
            <description>The column name is used to narrow down the search space. It will be added 
            to the filter condition. Please ensure it is encodable in `table_name`.</description>
        </column_name>
        <filter>
            <type>str</type>
            <required>false</required>
            <default></default>
            <description>The filter condition to narrow down the search space. Please refer to the 
            syntax of filter rules. By default, it is empty. It is suggested to restrict `pdf_id`,
            `page_number`, or `primary_key` to refine search results.</description>
        </filter>
        <limit>
            <type>int</type>
            <required>false</required>
            <default>5</default>
            <description>The number of top-ranked context to retrieve. Please ensure that it is a
            positive integer. And extremely large limit values may be truncated.</description>
        </limit>
    </parameters>
</action>
\end{Verbatim}
\textbf{\#\#\# Use Cases~(XML Format)}\newline\newline \ \textbf{\#\#\#\# Case 1}\newline Search the Mivlus collection \texttt{text\_bm25\_en}, which use BM25 sparse embeddings, with the filter condition ``\texttt{table\_name == 'chunks' and column\_name == 'text\_content' and pdf\_id == '12345678' and page\_number == 1}'' to restrict the content source and return the top $10$ relevant entries.

\textbf{[Action]:}
\begin{verbatim}
<action><action_type>RetrieveFromVectorstore</action_type><parameters><query>Does this paper discuss 
LLM-based agent on its first page?</query><collection_name>text_bm25_en</collection_name>
<table_name>chunks</table_name><column_name>text_content</column_name><filter>pdf_id == '12345678'
and page_number == 1</filter><limit>10</limit></parameters></action>
\end{verbatim}
\mbox{}\newline\textbf{\#\#\#\# Case 2}
\newline
.. more cases in XML format ...
\end{tcolorbox}

\begin{tcolorbox}[title={YAML Format}, colback=white,colframe=2,arc=1mm,boxrule=1pt,left=1mm,right=1mm,top=1mm,bottom=1mm]
\small
\color{2}
\textbf{\#\#\# Syntax and Parameters~(YAML Format)}
\begin{Verbatim}
action_type: RetrieveFromVectorstore
parameters:
    query:
        type: str
        required: true
        description: The query text will be encoded and used to search for relevant context. You
            can rephrase the user question to obtain a more clear and structured requirement.
    collection_name:
        type: str
        required: true
        description: The name of the collection in the Milvus vectorstore to search for relevant
            context. Please ensure the collection does exist in the vectorstore.
    table_name:
        type: str
        required: true
        description: The table name is used to narrow down the search space. It will be added to
            the filter condition. Please ensure this table has encodable columns.
    column_name:
        type: str
        required: true
        description: The column name is used to narrow down the search space. It will be added to
            the filter condition. Please ensure it is encodable in `table_name`.
    filter:
        type: str
        required: false
        default: ''
        description: The filter condition to narrow down the search space. Please refer to the
            syntax of filter rules. By default, it is empty. It is suggested to restrict `pdf_id`, 
            `page_number` or `primary_key` to refine search results.
    limit:
        type: int
        required: false
        default: 5
        description: The number of top-ranked context to retrieve. Please set it to
            a positive integer to limit the number of returned results. Extremely
            large limit values may be truncated.


\end{Verbatim}

\textbf{\#\#\# Use Cases~(YAML Format)}\newline\newline \ \textbf{\#\#\#\# Case 1}\newline Search the Mivlus collection \texttt{text\_bm25\_en}, which use BM25 sparse embeddings, with the filter condition ``\texttt{table\_name == 'chunks' and column\_name == 'text\_content' and pdf\_id == '12345678' and page\_number == 1}'' to restrict the content source and return the top $10$ relevant entries.

\textbf{[Action]:}
\begin{verbatim}
action_type: RetrieveFromVectorstore
parameters:
    query: Does this paper discuss LLM-based agent on its first page?
    collection_name: text_bm25_en
    table_name: chunks
    column_name: text_content
    filter: pdf_id == '12345678' and page_number == 1
    limit: 10
\end{verbatim}
\mbox{}\newline\textbf{\#\#\#\# Case 2}
\newline
.. more cases in YAML format ...
\end{tcolorbox}

\subsubsection{Observation Format}
\label{app:observation_format}
In this subsection, we discuss different observation formats to organize the retrieved entries. The execution result of action \retrievefromdatabase naturally forms a \emph{table}. While for action \retrievefromvectorstore, we extract the {\tt text} field for text type and the {\tt bounding\_box} field for image type respectively, and also concatenate the top-K entries as a \emph{table}. We support the following $4$ observation formats to serialize the table-stye observation: MARKDOWN, JSON, STRING, and HTML.

Take the \retrievefromdatabase action as an example, the SQL query to execute is:
\begin{center}
    {\tt select} {\it title}, {\it pub\_year} {\tt from} {\it metadata} {\tt where} {\it conference\_abbreviation} = {\tt 'ACL'} {\tt limit} $3$;
\end{center}
Then, the returned observation texts in different formats are:
\begin{tcolorbox}[title={Observation in MARKDOWN Format}, colback=white,colframe=6,arc=1mm,boxrule=1pt,left=1mm,right=1mm,top=1mm,bottom=1mm]
\small
\color{6}
\begin{verbatim}
+-----------------------------------------------------------+----------+
|                           title                           | pub_year |
+-----------------------------------------------------------+----------+
| ContraCLM: Contrastive Learning For Causal Language Model |   2023   |
|      Mitigating Label Biases for In-context Learning      |   2023   |
|    mCLIP: Multilingual CLIP via Cross-lingual Transfer    |   2023   |
+-----------------------------------------------------------+----------+
\end{verbatim}
In total, $3$ rows are displayed in MARKDOWN format.
\end{tcolorbox}


\begin{tcolorbox}[title={Observation in JSON Format}, colback=white,colframe=6,arc=1mm,boxrule=1pt,left=1mm,right=1mm,top=1mm,bottom=1mm]
\small
\color{6}
\begin{verbatim}
{"title":"ContraCLM: Contrastive Learning For Causal Language Model","pub_year":2023}
{"title":"Mitigating Label Biases for In-context Learning","pub_year":2023}
{"title":"mCLIP: Multilingual CLIP via Cross-lingual Transfer","pub_year":2023}
\end{verbatim}
In total, $3$ rows are displayed in JSON format.
\end{tcolorbox}

\begin{tcolorbox}[title={Observation in STRING Format}, colback=white,colframe=6,arc=1mm,boxrule=1pt,left=1mm,right=1mm,top=1mm,bottom=1mm]
\small
\color{6}
\begin{verbatim}
                                                    title  pub_year
ContraCLM: Contrastive Learning For Causal Language Model      2023
          Mitigating Label Biases for In-context Learning      2023
      mCLIP: Multilingual CLIP via Cross-lingual Transfer      2023
\end{verbatim}
In total, $3$ rows are displayed in STRING format.
\end{tcolorbox}

\begin{tcolorbox}[title={Observation in HTML Format}, colback=white,colframe=6,arc=1mm,boxrule=1pt,left=1mm,right=1mm,top=1mm,bottom=1mm]
\small
\color{6}
\begin{verbatim}
<table border="1" class="dataframe">
  <thead>
    <tr style="text-align: right;">
      <th>title</th>
      <th>pub_year</th>
    </tr>
  </thead>
  <tbody>
    <tr>
      <td>ContraCLM: Contrastive Learning For Causal Language Model</td>
      <td>2023</td>
    </tr>
    <tr>
      <td>Mitigating Label Biases for In-context Learning</td>
      <td>2023</td>
    </tr>
    <tr>
      <td>mCLIP: Multilingual CLIP via Cross-lingual Transfer</td>
      <td>2023</td>
    </tr>
  </tbody>
</table>
\end{verbatim}
In total, $3$ rows are displayed in HTML format.
\end{tcolorbox}

\subsection{Interaction Framework Prompt}
\label{app:interaction_prompt}
This section describes the interaction framework for iterative retrieval methods. For Classic and Two-stage RAG baselines, this part is omitted. We follow the popular ReAct framework~\cite{react} to encourage stepwise thought process before predicting actions.

\begin{tcolorbox}[title={Interaction Framework Prompt}, colback=white,colframe=4,arc=1mm,boxrule=1pt,left=1mm,right=1mm,top=1mm,bottom=1mm]
\small
\color{4}
\textbf{\#\# Interaction Framework}\newline
The main interaction procedure proceeds like this:\newline\newline\ - - - -\newline\newline[Thought]: reasoning process, why to take this action\newline[Action]: which action to take, please strictly conform to the action specification\newline[Observation]: execution results or error message after taking the action\newline\newline... more interleaved triplets of ([Thought], [Action], [Observation]) ...\newline\newline[Thought]: reasoning process to produce the final answer\newline[Action]: the terminal action \textasciigrave GenerateAnswer\textasciigrave, there is no further observation\newline\newline\ - - - -\newline\newline In general, the main interaction loop consists of an interleaved of triplets ([Thought], [Action], [Observation]), except the last \textasciigrave GenerateAnswer\textasciigrave action which does not have "[Observation]:". You need to predict the "[Thought]: ..." followed by the "[Action]: ..." for each turn, and we will execute your action in the environment and provide the "[Observation]: ..." for the previous action.
\end{tcolorbox}

\subsection{Hint Prompt}
This type of prompt provides a list of hints to guide the interaction. It highlights best practices for information retrieval, iterative refinement, and sequential decision-making throughout the task-solving process. It is highly customizable and can be easily extended to accommodate LLM prediction errors. Therefore, we only provide one demonstration example for the complete \ours framework.
\begin{tcolorbox}[title={Hint Prompt for \ours}, colback=white,colframe=5,arc=1mm,boxrule=1pt,left=1mm,right=1mm,top=1mm,bottom=1mm]
\small
\color{5}
\textbf{\#\# Suggestions or Hints for Agent Interaction}
\newline\newline
1. Explore multiple retrieval strategies. For example:

- Experiment with different (table, column) pairs to extract diverse types of information.

- Query various embedding models (collections) to find the most relevant context.
\newline\newline
2. Combine both structured and unstructured data. Concretely:

- Use SQL queries to retrieve precise facts and structured data. Pay special attention to morphological variations in cell values.

- Perform similarity searches in the vectorstore to capture semantic relationships and hidden insights.
\newline\newline
3. Iterate and refine:

- If SQL execution result is not satisfactory, try alternative SQL queries to explore the database content carefully.

- If the vector-based neural retrieval is insufficient, try alternative approaches or parameter settings.

- Use your findings to validate or enrich the final response.
\newline\newline
4. Ensure confidence. That is, only make a final decision when you are confident that the retrieved information fully addresses the user’s query.
\end{tcolorbox}

\subsection{Task Prompt}
The task prompt defines the concrete input for the current user question, which should at least include: 1) the user input question, 2) the answer format, and 3) the database or vectorstore schema~(if needed). Following \citet{nan2023enhancing}, we use the code representation format for database schema and incorporate schema descriptions to enhance the schema linking. As for the vectorstore schema, we introduce 1) all available collections, 2) fields for each stored data entry, 3) encodable (table, column) pairs from the corresponding DuckDB, and 4) valid operators to use in the {\tt filter} condition during vector search. We only present one input case for \ours, while the task prompt for other methods can be easily inferred depending on whether the database or vectorstore will be integrated as the backend environment.
\begin{tcolorbox}[title={Task Prompt for \ours}, colback=white,colframe=3,arc=1mm,boxrule=1pt,left=1mm,right=1mm,top=1mm,bottom=1mm]
\small
\color{3}
Remember that, for each question, you only have $20$ interaction turns at most. Now, let's start!
\newline
\textbf{[Question]:}  What are the main questions that this paper tries to resolve or answer?"
\newline
\textbf{[Answer Format]:} Your answer should be a Python list of text strings, with each element being one critical problem that this paper analyzes, e.g., ["question 1", "question 2", ...].
\newline
\textbf{[Database Schema]:} The database schema for ``{\tt ai\_research}'' is as follows:
\begin{verbatim}
/* database ai_research: This database contains information about AI research papers. Each PDF file is
    represented or parsed via different views, e.g., pages, sections, figures, tables, and references.
    We also extract the concrete content inside each concrete element via OCR. */
/* table metadata: This table stores metadata about each paper. */
CREATE TABLE IF NOT EXISTS metadata (
    paper_id UUID, -- A unique identifier for this paper.
    title VARCHAR, -- The title of this paper.
    abstract VARCHAR, -- The abstract of this paper.
    pub_year INTEGER, -- The year when this paper was published.
    ... [more columns with their data types and descriptions, omitted] ...
    PRIMARY KEY (paper_id)
);
... [the remaining tables and columns using the CREATE statement] ...
\end{verbatim}
\textbf{[Vectorstore Schema]:} The vectorstore schema for {\tt ai\_research} is as follows. You can try collections with different encoding models or modalities:
\begin{verbatim}
[
    {
        "collection_name": "text_bm25_en",
        "description": "This collection is used to store the sparse embeddings generated by the
            BM25 model for all encodable text content in another relational database. The semantic
            search is based on field `vector` with metric inner-product (IP).",
        "fields": [
            {"name": "vector", "dtype": "SPARSE_FLOAT_VECTOR", "desc": "attained by BM25 model"},
            {"name": "text", "dtype": "VARCHAR", "desc": "cell value from the database"},
            {"name": "pdf_id", "dtype": "VARCHAR", "desc": "unique id of the PDF file"},
            {"name": "page_number", "dtype": "INT16", "desc": "source page of the `text` field"},
            {"name": "table_name", "dtype": "VARCHAR", "desc": "source table of `text` field"},
            {"name": "column_name", "dtype": "VARCHAR", "desc": "source column of `text` field"},
            {"name": "primary_key", "dtype": "VARCHAR", "desc": "primary key value for the row
                that contains the `text` field in the relational database"}
        ]
    },
    {
        "collection_name": "text_sentence_transformers_all_minilm_l6_v2",
        "description": "This collection is used to store the embeddings generated by the model
            MiniLM-L6-v2 model for all encodable text content in another relational database. The
            semantic search is based on field `vector` with metric COSINE.",
        "fields": "The fields of this collection are the same as those in `text_bm25_en`."
    },
    ... [other collections with their fields] ...
]
\end{verbatim}
\mbox{}\newline
Here are all encodable (table\_name, column\_name) tuples from the corresponding DuckDB database, where the encoded vector entries are sourced. Different columns together provide multiple perspectives for vector search.
\begin{verbatim}
[("metadata", "title"), ("metadata", "abstract"), ... [more encodable (table, column) pairs] ...]
\end{verbatim}
\mbox{}\newline
Here are the operators that you can use in the {\tt filter} parameter for RetrieveFromVectorstore action:
\begin{verbatim}
[
    {
        "symbol": "and",
        "example": "expr1 and expr2",
        "description": "True if both expr1 and expr2 are true."
    },
    {
        "symbol": "+",
        "example": "a + b",
        "description": "Add the two operands."
    },
    ... [more operators with detailed description] ...
]
\end{verbatim}
\end{tcolorbox}
\end{document}